%% file: Survey.tex
\documentclass[acmlarge]{acmart}
\usepackage{bm}    
\usepackage{multirow}

\usepackage{amssymb}

\usepackage{amsmath}
\usepackage{colortbl}
\usepackage{xcolor} 
\usepackage{forest}        
\usepackage{tikz}
\usepackage{graphicx}
\usepackage{tcolorbox}
\usepackage{array}
\usepackage{hyperref}
\usepackage{booktabs}
\usepackage{multirow}
\usepackage{cleveref}
\usepackage{xspace}  
\usepackage{booktabs}
\usepackage{wrapfig}
\usepackage{colortbl}
\usepackage{threeparttable}
\usepackage{caption} 
\usepackage{todonotes}
\usetikzlibrary{trees}
\usetikzlibrary{shapes.geometric, arrows}
\AtBeginDocument{%
  }
\newcommand{\seclink}[1]{\hyperref[#1]{\textsection}}

\setcopyright{acmlicensed}
\copyrightyear{2018}
\acmYear{2018}
\acmDOI{XXXXXXX.XXXXXXX}

\acmJournal{POMACS}
\acmVolume{37}
\acmNumber{4}
\acmArticle{111}
\acmMonth{8}

\begin{document}

\title{Large Language Models Meet Text-Attributed Graphs: A Survey of Integration Frameworks and Applications}


\author{Guangxin Su}
\affiliation{%
  \institution{The University of New South Wales}
  \city{Sydney}
  \country{Australia}}
\email{guangxin.su@unsw.edu.au}

\author{Hanchen Wang}
\affiliation{%
  \institution{The University of Technology Sydney}
  \city{Sydney}
  \country{Australia}}
\email{hanchen.wang@uts.edu.au}

\author{Jianwei Wang}
\affiliation{%
  \institution{The University of New South Wales}
  \city{Sydney}
  \country{Australia}}
\email{jianwei.wang1@unsw.edu.au}

\author{Wenjie Zhang}
\affiliation{%
  \institution{The University of New South Wales}
  \city{Sydney}
  \country{Australia}}
\email{wenjie.zhang@unsw.edu.au}

\author{Ying Zhang}
\affiliation{%
  \institution{University of Technology Sydney}
  \city{Sydney}
  \country{Australia}}
\email{ying.zhang@uts.edu.au}


\author{Jian Pei}
\affiliation{%
  \institution{Duke University}
  \city{Durham}
  \country{USA}}
\email{j.pei@duke.edu}







\renewcommand{\shortauthors}{Su et al.}

\begin{abstract}

{Large Language Models (LLMs) have achieved remarkable success in natural language processing through strong semantic understanding and generation. However, their black-box nature limits structured and multi-hop reasoning. In contrast, Text-Attributed Graphs (TAGs) provide explicit relational structures enriched with textual context, yet often lack semantic depth. Recent research shows that combining LLMs and TAGs yields complementary benefits: enhancing TAG representation learning and improving the reasoning and interpretability of LLMs.
This survey provides the first systematic review of LLM–TAG integration from an orchestration perspective. We introduce a novel taxonomy covering two fundamental directions: LLM for TAG, where LLMs enrich graph-based tasks, and TAG for LLM, where structured graphs improve LLM reasoning. We categorize orchestration strategies into sequential, parallel, and multi-module frameworks, and discuss advances in TAG-specific pretraining, prompting, and parameter-efficient fine-tuning. Beyond methodology, we summarize empirical insights, curate available datasets, and highlight diverse applications across recommendation systems, biomedical analysis, and knowledge-intensive question answering. Finally, we outline open challenges and promising research directions, aiming to guide future work at the intersection of language and graph learning.}

\end{abstract}


\ccsdesc[500]{Computing methodologies~Natural language processing; Knowledge representation and reasoning}
\ccsdesc[500]{Mathematics of computing~Graph
algorithms}
\ccsdesc[300]{General
and reference~Surveys and overviews}

\keywords{Large Language Models, Text-attributed Graphs, Graph Neural Networks, Natural
Language Processing, Graph Representation Learning}

\received{20 February 2007}
\received[revised]{12 March 2009}
\received[accepted]{5 June 2009}

\maketitle

\input{s1-introduction}

\input{s2-prelinminary}

\input{s3-roadmap}
\input{s4-llm4tag}
\input{s5-tag2llm}
\input{s6-direction}

\input{s7-temp-insights}

\bibliographystyle{ACM-Reference-Format}


\end{document}

%% file: s1-introduction.tex









\section{INTRODUCTION}
\begin{wrapfigure}{r}{8cm}
\centering
\includegraphics[width=0.49\textwidth]{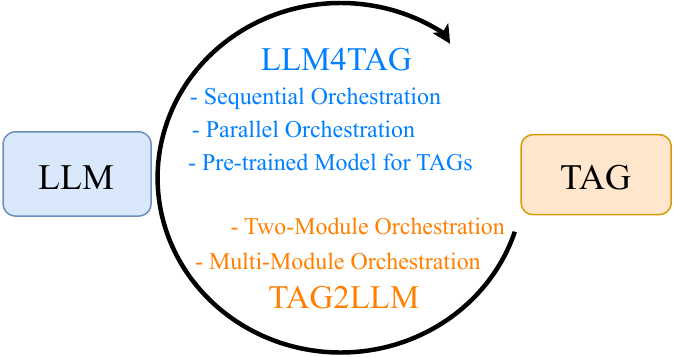}
\caption{\footnotesize Synergistic improvement with the orchestration of techniques for large language models and text-attributed graphs.}
\vspace{-3mm}
\label{fig:advantage}
\end{wrapfigure}

The rise of Transformer-based architectures \cite{vaswani2017attention} has revolutionized natural language processing (NLP). 
Exemplifying large language models (LLMs), the GPT series \cite{radford2018improving, radford2019language, brown2020language, achiam2023gpt}, built on the Transformer decoder architecture and pre-trained on extensive corpora, has demonstrated extraordinary capabilities, positioning LLMs as foundational steps toward realizing Artificial General Intelligence (AGI). 
In contrast, more compact language models (LMs\footnote{The comparison between LMs and LLMs highlights a trade-off between capability and flexibility. LLMs, encompassing LMs with their larger parameter sizes, offer greater capabilities and extensive knowledge but often sacrifice fine-tuning efficiency. In contrast, the more compact LMs excel in adaptability for specific tasks \cite{he2023harnessing, chen2024exploring}.}) such as the BERT series \cite{devlin2018bert, he2020deberta, liu2019roberta, sanh2019distilbert}, leverage the Transformer encoder to excel in tasks demanding fine-grained contextual understanding and precise semantic representation, making them essential for domain-specific research and applications.

In the real world, a vast amount of critical information, ranging from scientific publications and social media posts to biological records, is stored in textual form \cite{stelzer2016genecards, hu2020open, li2024empowering}. 
These textual data often exhibit rich interrelationships, which can be naturally modeled as text-attributed graphs (TAGs)~\cite{yang2021graphformers}, where nodes, edges, or the entire graph are enriched with textual attributes.
Therefore, there are two data types within TAGs, \textit{textual attributes} and \textit{graph structure}.
For instance, in citation networks, nodes represent research papers annotated with abstracts or full texts, while edges denote citation links that reflect semantic relevance. 
In recommendation systems, TAGs can represent users and items, where nodes are enriched with user profiles or product descriptions, and edges capture user-item interactions, co-purchase behaviors, or review-based sentiment information.
In chemistry studies, chemical compounds can be modeled as TAGs, where atoms or entire molecules are annotated with textual property descriptions or literature-derived knowledge. 
Accurately modeling and analyzing these structured yet text-rich relationships is crucial for a variety of downstream tasks, including text classification \cite{zhao2022learning}, personalized recommendation \cite{zhu2021textgnn}, and molecular discovery \cite{jablonka202314}.

Recently, growing attention has been directed towards orchestrating LLM and TAG techniques, driven by their combined ability to handle textual and graph-structured data.
LLMs excel in parsing and generating text, while TAGs capture complex relational structures inherent in data.
Figure \ref{fig:advantage} demonstrates diverse orchestration strategies for LLM and TAG techniques in handling textual and graph-structural data.
These integrated approaches not only improve performance on TAG-oriented tasks (\textbf{LLM for TAGs}) but also enhance the reasoning capabilities of LLMs (\textbf{TAG for LLMs}), which have yielded promising advances across a range of research domains, including graph retrieval-augmented generation \cite{peng2024graph}, knowledge-intensive question answering \cite{pan2024unifying}, the mitigation of hallucinations in LLMs \cite{guan2024mitigating}, and scientific discovery \cite{shaoastronomical, gu2024empowering}, among others.
For example, \cite{yue2020graph} constructs a TAG from electronic medical records (EMRs).
Medical terms are represented as nodes, and edges are formed based on co-occurrence within context windows \cite{finlayson2014building}. 
The integration of LLM and TAG techniques enhances both semantic and structural representations, enabling more accurate and interpretable classification of medical terms.
Motivated by increasing research interest in the integration of LLM and TAG techniques, this paper presents a comprehensive survey from a novel perspective: \textit{how these techniques are orchestrated to provide synergistic improvements for both the reasoning ability of LLMs and representation learning of TAGs}. 
We systematically organize existing works through two fundamental primitives, LLM4TAG and TAG4LLM, introducing a novel taxonomy of orchestration frameworks.
\begin{enumerate}
\item[$\star$] \noindent \textbf{LLM for TAG. }
Applying LLMs to textual attributes in TAGs produces context-rich embeddings that, when combined with topology-aware graph learning mechanisms, yield superior representation learning of TAGs. 
In a \textbf{sequential orchestration}~\cite{he2023harnessing} scheme, the LLM first encodes textual features, and its output is then passed to a graph learning model that explicitly models connectivity. 
In a \textbf{parallel orchestration}~\cite{feng2021zero} scheme, the LLM and graph learning model run in tandem, each processing text and topology respectively, before aligning their embeddings for downstream tasks using techniques such as contrastive learning.
Moreover, several works~\cite{bielak2022graph, wu2021self} repurposed successful LLM training paradigms, including self-supervised pretraining and fine tuning, to strengthen TAG learning. 
In particular, \textbf{pre-trained models for TAGs}~\cite{xie2023graph} adapt self-supervised objectives, fine-tuning routines, and prompt design strategies from LLMs into the graph domain, creating more expressive and generalizable TAG frameworks.

\item[$\star$] \noindent \textbf{TAG for LLM.}
TAGs, with their structured and explicit integration of textual attributes, provide a clear substrate for symbolic reasoning that helps LLMs address transparency and decisiveness challenges. This issue arises because LLMs embed vast knowledge implicitly within their parameters \cite{zhang2023survey}, leading to opacity that hinders interpretability and factual precision. Techniques like chain-of-thought prompting \cite{wei2022chain} attempt to generate explanations but still suffer from hallucinations \cite{ji2023survey} and inaccuracies \cite{wang2023survey}, especially in multi-hop reasoning. By grounding LLMs in TAGs, it becomes possible to produce reliable, interpretable outputs and mitigate these limitations \cite{pan2024unifying}.
We systematically investigate TAG for LLM techniques and divide them into two categories: \textbf{two-module orchestration}~\cite{chen2024llaga}, and \textbf{multi-module orchestration}~\cite{gao2023retrieval}.
In both cases, the underlying principle is to combine symbolic, topology-aware information generated from TAGs with the LLM’s language capabilities, yielding outputs that are more transparent, interpretable, and factually precise.
\end{enumerate}

\noindent
\textbf{Distinction with Existing LLM-TAGs Surveys.}
This is the first comprehensive survey that summarizes the LLM and TAG models from the perspective of \textit{model orchestration and mutual enhancement}, i.e., how the data and techniques are organized and utilized in recent works.
In this article, we provide a comprehensive overview of how techniques from LLM and TAG research areas are orchestrated for the improvement of models for LLMs and TAGs. 
These methods aim to refine the reasoning abilities of LLMs while enhancing the effectiveness of TAG representation learning. 
\textit{Scope Expansion of LLM-TAG Techniques: }
Our survey goes beyond prior formal surveys by systematically covering every detail about LLM4TAG and TAG4LLM, 
which are often missing or only partially addressed in existing surveys. 
Additionally, we include pre-training methods, including TAG-based self-supervised learning and transformer-based models, providing a broader and deeper exploration of the synergies between LLMs and TAGs, advancing towards a foundational model for graphs.
This survey also provides the techniques for applications on \textit{multiple levels} of graphs.
Previous surveys \cite{jin2024large, li2023survey, ren2024survey, chen2024exploring, fan2024graph, mao2024advancing} on TAGs primarily focus on the node level, treating text as node attributes. 
Our survey expands TAGs to edge and graph levels, emphasizing how textual information enhances relationships (edges) and structured knowledge (graphs, especially KGs \cite{pan2024unifying, peng2024graph}). 
We analyze shared techniques and key differences across these levels, offering a holistic view of LLM-TAG integration beyond node-centric approaches.
\textit{Thorough Summarization of Experimental Observations and Insights:}
Among existing surveys, only  \cite{chen2024exploring} provides the summary of the observations and insights from the experiment results, but their analysis remains limited to a narrow set of existing models \cite{zhao2022learning, he2023harnessing}. 
In contrast, we systematically synthesize insights from recent, relevant studies, providing a more structured and comprehensive perspective. 
Moreover, we offer key observations and empirical insights, establishing a foundation to guide future research.
Furthermore, we extend beyond homogeneous TAGs by incorporating multi-modal TAGs, which have been largely missed in previous surveys. 
Additionally, we provide a comprehensive overview of real-world applications, broadening the scope of research. 
To further support future studies, we systematically collect and curate relevant datasets, offering a valuable resource for advancing research about LLM and TAG.

\noindent
\textbf{\textit{Contributions:}}
The contributions of this survey can be summarized as follows:
\begin{enumerate}
    \item[(1)] \textbf{Novel perspective for holistic survey.}
    This survey presents the existing works from a novel perspective: how the techniques and data are orchestrated to provide the synergistic improvements for LLM and TAG.
    \item[(2)] \textbf{Techniques categorization.} 
    We comprehensively review and categorize how LLMs boost the performance of TAGs, along with how TAGs enhance LLMs, focusing on key strategies and underlying concepts to provide vital perspectives on each framework. 
    \item[(3)] \textbf{Experimental observations and insights summarization. }
    We summarize our observations from the experimental results of existing studies, synthesizing insights across three dimensions: recent advancements, challenges and limitations, and potential avenues for further development.
    \item[(4)] \textbf{Resources and future directions.} We curate comprehensive datasets and outline future research directions, spanning from data management to unexplored architectural innovations for LLMs and TAGs.
\end{enumerate}

\noindent
\textbf{\textit{Organization: }}
The structure of this paper is organized as follows:
\Cref{sec:defination} introduces the background and preliminaries of this paper.
\Cref{sec:llm4tag} surveys the technical frameworks, real-world applications, and empirical insights on how LLMs are orchestrated within TAG modeling. 
In turn, \Cref{sec:tag2llm} reviews the frameworks, real-world applications, and observed insights on how TAGs are orchestrated within LLM pipelines to strengthen reasoning. \Cref{sec:futureDire} discusses open challenges and opportunities. 
Finally, \Cref{sec:conclusion} concludes the survey.


\begin{figure}[t]  
\centering
%
\vspace{-2mm}
\resizebox{\linewidth}{!}{
\begin{forest}
for tree={
  grow=east,
  draw,
  edge=ultra thick,
  parent anchor=east,
  child anchor=west,
  edge path={
    \noexpand\path [\forestoption{edge}] 
      (!u.parent anchor) -- +(8pt,0) |- (.child anchor)\forestoption{edge label};
  },
  anchor=west,
  l sep+=8pt,
  s sep+=2pt,
  font=\scriptsize,
  calign=center
},
where level=0{ text width=3.0cm }{},
where level=1{ text width=2.2cm }{},
where level=2{ text width=3.0cm }{},
where level=3{ text width=4.5cm }{},
[Large Language Models and Text-Attributed Graphs
  [{\seclink{sec:tag2llm} TAG for LLM}, edge=black
    [\seclink{4.4obser} Observations and Insights
    ]
    [\seclink{4.3applicaiton} Applications
    ]
    [\seclink{4.2} Multi-Module Orchestration
      [\seclink{subsec:tog} Graph of Thought]
      [\seclink{subsec:RAG} Retrieval-Augmented Knowledge]
      [\seclink{subsec:nested} Nested Structure]
    ]
    [\seclink{4.1} Two-Module Orchestration
      [\seclink{subsub:softPrompt} Soft Prompt-based Fusion]
      [\seclink{subsec:embedLLM} Embedding-based Fusion]
    ]
  ]
  [\seclink{sec:llm4tag} LLM for TAG, edge=black
    [\seclink{subsec:3.5obser} Observations and Insights
    ]
    [\seclink{subsec:3.4application} Applications
    ]
    [\seclink{subsec:pretrainTAG} Pre-trained Model for TAGs
      [\seclink{subsec:SSLprompt} Graph Prompt Tuning]
      [\seclink{subsec:SSLft} Fine-tuning]
      [\seclink{subsec:SSL} Graph Foundational Models]
      [\seclink{subsec:graphTransformer} TAG Transformers]
    ]
    [\seclink{subsec:parallel} Parallel Orchestration
      [\seclink{subsub:mutualPseduoL} Mutual Supervision via Pseudo-Labels]
      [\seclink{subsub:2tower} Two-tower Structure]
    ]
    [\seclink{subsec:3.1} Sequential Orchestration
      [\seclink{subsub:MoE} Mixture of Experts]
      [\seclink{subsub:structureLM} Structure-aware Language Models]
      [\seclink{subsubstr-prompt} Structure-aware Prompts]
      [\seclink{subsec:AgnosticPrompt} Structure-agnostic Prompts]
      [\seclink{subsec:3.1.1} Prompting on TAGs]
    ]
  ]
]
\end{forest}}
\vspace{-2mm}  
\caption{Holistic categorisation of approaches for Large Language Models and Text-Attributed Graphs.}
\label{fig:catalyst-taxonomy}
\end{figure}
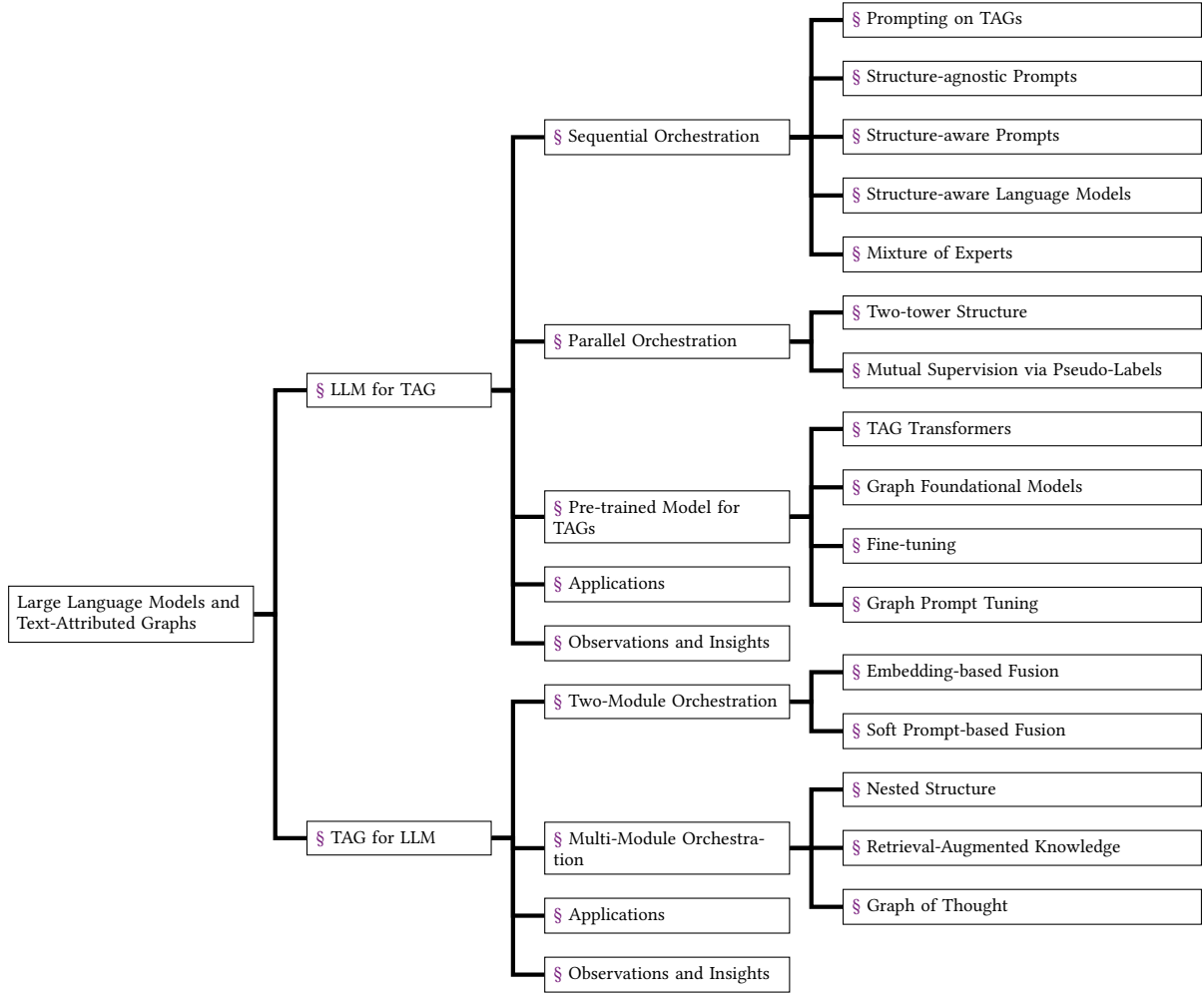

%% file: s2-prelinminary.tex
\section{BACKGROUND \& PRELIMINARIES}
\label{sec:defination}




\subsection{Background}

A \textbf{text-attributed graph}~\cite{yang2021graphformers} is a graph structure that incorporates textual attributes at the node level, edge level, or graph level, and is defined as:
\(
\mathcal{G} = (\mathcal{V}, \mathcal{E}, \bm{T_\mathcal{V}}, \bm{T_\mathcal{E}}),
\). 
Here, \(\mathcal{V}\) is the set of nodes, and \(\mathcal{E} \subseteq \mathcal{V} \times \mathcal{V}\) is the set of edges in the graph. 
The set \(\bm{T_\mathcal{V}} = \{\bm{t}_v\}_{v \in \mathcal{V}}\) (\textit{resp}. \(\bm{T_\mathcal{E}} = \{\bm{t}_e\}_{e \in \mathcal{E}}\) ) contains the textual attributes at the node level (\textit{resp}. edge level), where each \(\bm{t}_{v_i} \in \mathcal{D}^{L_{v_i}}\) (\textit{resp}. \(\bm{t}_{e_i} \in \mathcal{D}^{L_{e_i}}\)) is a token sequence.
The \(\bm{T_\mathcal{G}}\) is the global textual attribute associated with the entire graph.
A text-attributed graph is referred to as a node-level TAG, edge-level TAG, or graph-level TAG when textual attributes are present exclusively at the node level, edge level, or graph level, respectively.

TAGs contain rich structural and textual information, enabling them to more comprehensively represent real-world entities and their complex relationships.
Meanwhile, LLMs, with billions of parameters, have exhibited remarkable emergent behaviors. They have successfully handled diverse data types such as images~\cite{li2023blip, liu2024improved}, textual~\cite{chen2024llaga}, and tabular~\cite{jin2025hegta}, and demonstrated impressive zero-shot generalization across a wide range of tasks.
Despite these advancements, orchestrating LLMs and TAGs to meet various application needs remains an open and challenging problem. On one hand, the richness and heterogeneity of structural and textual information in TAGs pose significant challenges for LLMs to effectively process and reason over them. On the other hand, how to infuse structured knowledge from TAGs into LLMs to improve their reasoning abilities, factual consistency, and adaptability is still largely unexplored.



%

In this survey, we explore the orchestration of LLMs and TAGs to address a wide range of real-world applications.
In particular, we focus on two fundamental primitives, namely \textbf{LLM for TAG} and \textbf{TAG for LLM}, which serve as the building blocks for developing diverse task-centric orchestrations.

LLM for TAG refers to the pipeline that focuses on applying LLM techniques to improve performance on TAG-related tasks. Given a TAG as input, various LLM-centric strategies such as prompt design, instruction tuning, or in-context learning are employed to enhance the ability of the LLM to understand and reason over the graph structure and textual attributes. 
This pipeline enables LLMs to act as powerful tools for processing and analyzing complex TAG data.


TAG for LLM is a pipeline that focuses on leveraging TAGs to enhance the capabilities of LLMs.
It investigates how structured knowledge from TAGs can be integrated into LLMs to improve their reasoning, factual consistency, and adaptability.
This pipeline injects external structured knowledge from TAGs into LLMs, thereby augmenting their performance on knowledge-intensive tasks such as question answering, grounded text generation, and retrieval-augmented inference.

\subsection{Preliminaries}

TAGs contain rich structural and textual information. To capture structural patterns, graph learning models are commonly used to aggregate information based on connectivity. For textual attributes, pre-trained language models are typically employed to extract semantic representations at the node, edge, or graph level.
In this subsection, we introduce key techniques relevant to this survey, focusing in particular on graph learning models and techniques for language models.

\subsubsection{Graph Learning Models}


There is a long history of research on graph learning models in the literature. In the early stages, attention was primarily focused on message-passing graph neural networks (GNNs). In a typical message-passing GNN, the update of the representation of a node can be expressed as:

\[
h_v^{(k)} = \text{Update}^{(k)} \left( h_v^{(k-1)}, \{ \text{Aggregate}^{(k)}(h_u^{(k-1)}, e_{vu}) | u \in \mathcal{N}(v) \} \right)
\]
Where \( h_v^{(k)} \) is the representation of node \( v \) at the \( k \)-th iteration. \( \mathcal{N}(v) \) denotes the set of neighbors of node \( v \). \( h_u^{(k-1)} \) is the representation of neighbor node \( u \) in the previous iteration. \( e_{vu} \) represents the edge information between nodes \( v \) and \( u \) (if any). \( \text{Aggregate}^{(k)} \) is the aggregation function, which typically combines the information from the neighbors. \( \text{Update}^{(k)} \) is the update function, which combines the previous node representation and the aggregated neighbor information to produce the new node representation.
Models like GCN \cite{kipf2016semi} and GraphSAGE \cite{hamilton2017inductive} are efficient for large-scale semi-supervised learning, where GCN focuses on simple aggregation through weighted sums, while GraphSAGE introduces more flexible aggregation strategies, such as mean, pooling, and Long Short-Term Memory (LSTM). GAT \cite{velivckovic2017graph} enhances aggregation with attention mechanisms, allowing the model to dynamically weight the importance of neighboring nodes. RGCN \cite{schlichtkrull2018modeling} extends aggregation by handling multi-relational graphs, incorporating edge types into the message-passing process. Advanced models like GIN \cite{xu2018powerful} improve the update step by using a more expressive aggregation function, where a multi-layer perceptron (MLP) is applied to better capture complex graph structures.


Besides the traditional message-passing GNN, new advanced architectures, such as graph transformers~\cite{shehzad2024graph, rampavsek2022recipe} and graph mamba~\cite{behrouz2024graph,wang2024graph,li2024stg},
are emerging in the literature to further enhance the graph learning performance.
Graph transformers \cite{shehzad2024graph, rampavsek2022recipe} adapt the transformer architecture to graph-structured data by using self-attention to capture long-range dependencies between nodes, enhancing their ability to model complex graph structures beyond local neighborhoods.
Compared to traditional GNNs, it offers greater expressive power while preserving structural information, making it effective for tasks like node representation and graph classification. 
Another representative architecture is the graph mamba, which is a state-space-based graph learning model that replaces traditional message passing with parallelizable sequence modeling, enabling long-range dependency capture across graph structures. 
Numerous graph mamba models have been developed for various graph data types, including spatial-temporal graphs~\cite{li2024stg} and dynamic graphs~\cite{li2024dyg,yuan2025dg}.

Graph learning can be broadly categorized into supervised, semi-supervised, and unsupervised paradigms.
Supervised methods such as GIN and GraphSAGE rely on full labels for tasks like molecular property prediction and graph classification. Semi-supervised methods like GCN and GAT use a small set of labeled nodes and graph structure to infer labels for the rest.  In unsupervised graph learning, methods like clustering (e.g., spectral clustering~\cite{berahmand2025comprehensive}), matrix factorization (e.g., probabilistic matrix factorization~\cite{li2025revisiting}), or random walk (e.g., DeepWalk~\cite{bozorgi2025survey}) aim to learn node or graph representations without relying on labels.
Self-supervised methods represent an important mainstream approach within unsupervised learning.
Graph self-supervised learning models, such as GraphCL \cite{you2020graph}, leverage unlabeled graph structures to improve representation learning, model performance, and generalization.
GraphCL adopts a contrastive learning framework by generating multiple augmented views of the same graph through perturbation strategies, including node dropping, edge perturbation, attribute masking, and subgraph sampling.
The learned representations are used for downstream tasks through fine-tuning \cite{lu2021learning} or graph prompt tuning \cite{sun2022gppt, sun2023all} strategies.

\begin{table*}[h!]
\scriptsize
\caption{LLMs used in TAG research, grouped by Architecture and Domain.}
\label{TAB:llmClassifiedDetailed}
\resizebox{\textwidth}{!}{
\begin{tabular}{lllccc}
\toprule
\textbf{Architecture} & \textbf{Domain} & \textbf{LLM} & \textbf{\# Parameters} & \textbf{Year} & \textbf{Brief Description} \\
\midrule
\multirow{17}{*}{\textbf{Decoder-Only}} 
    & \multirow{14}{*}{General-purpose} 
        & DeepSeek-R1~\cite{guo2025deepseek} & 1.5B–70B & 2025 & RL-trained model with strong reasoning ability.\\
    &  & LLaMA-3~\cite{dubey2024llama} & 8B–405B & 2024 & Multilingual, coding, reasoning, tool use.\\
    &  & Mixtral 8×7B~\cite{jiang2024mixtral} & 56B & 2024 & Sparse MoE (8×LLaMA-7B) for efficiency.\\
    &  & Claude 3 Haiku & – & 2024 & Tuned for speed and cost-effectiveness.\\
    &  & Claude 3 Sonnet & – & 2024 & Balanced intelligence–latency trade-off.\\
    &  & Vicuna~\cite{chiang2023vicuna} & 7B, 13B & 2023 & Chatbot fine-tuned on ShareGPT data.\\
    &  & LLaMA-2~\cite{touvron2023llama} & 7B–70B & 2023 & Base and conversational variants.\\
    &  & LLaMA-2-chat~\cite{touvron2023llama} & 7B–70B & 2023 & Chat-optimised LLaMA-2.\\
    &  & GPT-4~\cite{achiam2023gpt} & – & 2023 & Large multimodal model (text/image).\\
    &  & GPT-3.5~\cite{brown2020language} & – & 2020 & Enhanced comprehension and generation.\\
    &  & PaLM~\cite{chowdhery2023palm} & 540B & 2022 & Scalable, strong multilingual reasoning.\\
    &  & Mistral 7B~\cite{jiang2023clip} & 7B & 2024 & Dense, efficient open-source baseline.\\
    &  & Qwen 3~\cite{yang2025qwen3} & 0.5B–72B & 2025 & Chinese–English LLMs with instruction tuning.\\
    &  & Baichuan 2~\cite{yang2023baichuan} & 7B, 13B & 2023 & High-quality bilingual open models.\\
\cline{2-6}
    & \multirow{2}{*}{Multimodal} 
        & GPT-4V~\cite{2023GPT4VisionSC} & – & 2023 & Vision-augmented GPT-4; image + text input.\\
    &  & Gemini~\cite{team2023gemini} & 1.8B–1T & 2023 & Handles image, audio, video and text.\\
\cline{2-6}
    & Scientific 
        & Galactica~\cite{taylor2022galactica} & 1.3B & 2022 & Stores and reasons over scientific knowledge.\\
\midrule
\multirow{10}{*}{\textbf{Encoder-Only}} 
    & \multirow{4}{*}{General-purpose} 
        & BERT~\cite{devlin2018bert} & 110M, 340M & 2018 & Contextualised word embeddings.\\
    &  & RoBERTa~\cite{liu2019roberta} & 125M, 355M & 2019 & Robustly optimised BERT.\\
    &  & DistilRoBERTa & 82.8M & 2020 & Lightweight distilled RoBERTa.\\
    &  & DeBERTa~\cite{he2020deberta} & 140M–1.6B & 2020 & Disentangled attention.\\
\cline{2-6}
    & \multirow{3}{*}{Retrieval-oriented} 
        & Sent-BERT~\cite{reimers2019sentence} & 22M & 2019 & Sentence embeddings for similarity search.\\
    &  & e5-large-v1~\cite{wang2022text} & 560M & 2022 & Strong retrieval / clustering embeddings.\\
    &  & gte-Qwen1.5-7B-inst.~\cite{li2023towards} & 7B & 2023 & Instruction-tuned gte model.\\
\cline{2-6}
    & Biomedical 
        & BioBERT~\cite{lee2020biobert} & 110M & 2020 & Pre-trained on biomedical corpora.\\
\cline{2-6}
    & Molecular 
        & ChemBERTa~\cite{chithrananda2020chemberta} & 77M & 2020 & Embeds SMILES strings.\\
\cline{2-6}
    & Scientific 
        & SciBERT~\cite{beltagy2019scibert} & 110M & 2019 & Pre-trained on scientific texts.\\
\midrule
\multirow{2}{*}{\textbf{Encoder–Decoder}} 
    & General-purpose 
        & FLAN-T5~\cite{chung2024scaling} & 80M–11B & 2024 & Instruction-tuned T5.\\
\cline{2-6}
    & Molecular 
        & MolT5~\cite{edwards-etal-2022-translation} & 77M–880M & 2022 & Joint NL + molecule pre-training.\\
\bottomrule
\end{tabular}}
\end{table*}

\subsubsection{Language Models}
A language model, which is a machine learning model typically constructed using efficient Transformer variants, is designed to understand and generate human language by predicting the likelihood of word sequences.
Language models vary in size and capability. 
Representative language models are summarized in Table \ref{TAB:llmClassifiedDetailed}. 
In the early stage, the research attention mainly focus on the relatively small language model ($e.g., $BERT \cite{devlin2018bert}).
The models are designed to be lightweight neural language models, typically ranging from a few million to a few hundred million, enabling fast inference and easy deployment on resource-constrained environments. 
They are primarily based on encoder-decoder or encoder-only architectures.
Their effectiveness largely comes from pretraining on large-scale unlabeled corpora using masked language modeling (MLM) or next-sentence prediction (NSP), followed by fine-tuning on downstream tasks.
Models like BERT and Sent-BERT \cite{reimers2019sentence} provide robust contextual embeddings, ensuring practical viability and superior performance in resource-constrained settings. Enhanced variants like DeBERTa \cite{he2020deberta} with disentangled attention and RoBERTa \cite{liu2019roberta} with enhanced pretraining strategies yield strong contextual representations and are well suited for TAG tasks requiring fine granularity.
Additionally, domain-specific adaptations, such as BioBERT \cite{lee2020biobert} and SciBERT \cite{beltagy2019scibert}, cater to specialized scientific and biomedical contexts.

As motivated by the scaling law~\cite{achiam2023gpt}, which demonstrates that model capacity increases with the number of parameters, recent research attention has shifted toward LLM, focusing on how scaling up model size and complexity leads to significant improvements in performance across a wide range of textual tasks~\cite{yang2024harnessing, naveed2023comprehensive, pan2024unifying}.
Unlike SLMs, LLMs typically contain billions of parameters and are usually auto-regressive, built upon a decoder-only architecture. They are trained on massive text corpora by maximizing the log-likelihood of the next token, given the preceding context:

\begin{equation} \theta_{LLM} = \mathop{\arg\max}\limits_{\theta} \sum\limits_{i} \log P(t_i \mid t_{i-k}, \cdots, t_{i-1}; \theta) \end{equation}
where $t_i$ denotes the $i$-th token and $k$ represents the context window size.
\noindent
A majority of LLMs fall under the category of foundation models, which serve as a general-purpose backbone to support zero-shot and few-shot learning across diverse domains.
Representative models like GPT-4 \cite{achiam2023gpt}, DeepSeek-V3\cite{guo2025deepseek}, and LLaMA-3 \cite{dubey2024llama} excel in multilinguality and tool integration, making them ideal for high-performance and complex requirements. 
Open-source solutions such as Vicuna \cite{chiang2023vicuna}, optimized for prompt engineering and fine-tuning, offer flexible conversational capabilities. 
Sparse models like Mixtral~\cite{jiang2024mixtral}, utilizing a mixture-of-experts framework, enable dynamic adaptability to diverse tasks and input complexities.
Besides foundation models, recent research has introduced specialized reasoning models that aim to enhance the logical inference and multi-step reasoning capabilities of LLMs. 


There are two typical ways to adapt the model to new tasks and boost performance: fine-tuning and prompting.

\noindent\textbf{Fine-tuning}.
\noindent
Fine-tuning remains the dominant way to adapt LLMs to downstream graph tasks, yet full-parameter updates are infeasible for billion-scale backbones; consequently, recent work on TAGs has embraced parameter-efficient fine-tuning (PEFT). 
Techniques such as LoRA \cite{hu2022lora}, which inserts low-rank adapters into attention matrices, Prefix/Prompt-Tuning \cite{lester2021power, li2021prefix}, which optimises a virtual token sequence at every layer, IA$^{3}$, which scales key–value and feed-forward channels with tiny learned vectors, and AdapterFusion \cite{pfeiffer2021adapterfusion}, which learns task-specific gating over a bank of lightweight adapters, can all be trained on commodity GPUs while matching or exceeding full fine-tuning accuracy on node classification and graph retrieval benchmarks \cite{hu2022lora,li2021prefix,lester2021power}. 
PEFT has already boosted state-of-the-art TAG models such as GraphLoRA \cite{10.1145/3690624.3709186}. 
Crucially, pure GNNs alone struggle with TAGs because they excel at propagating structural signals yet lack the capacity to capture the semantic compositionality of free text, require large labelled corpora to learn textual representations from scratch, and cannot leverage the encyclopaedic world knowledge embedded in modern LLM pre-training; integrating LLMs with TAG-aware GNNs therefore marries rich linguistic context with relational inductive biases, yielding models that remain robust under label sparsity and distribution shift.

\noindent\textbf{Prompting}.
\noindent
Prompt engineering \cite{wei2022chain, gao2020making, liu2023pre} is a versatile method to guide LLMs ($e.g.,$ GPT-4) by specifying tasks through carefully crafted input prompts without altering the model's parameters.
It includes zero-shot prompting, where models rely solely on pre-trained knowledge to handle tasks without labeled examples \cite{radford2019language}, and few-shot prompting, which provides a few task-specific examples to improve performance \cite{brown2020language}. 
While zero-shot prompting eliminates the need for training data, few-shot prompting enhances capabilities for complex tasks but requires careful selection of examples and additional input tokens \cite{sahoo2024systematic}.
By leveraging task-specific instructions, few-shot examples, or structured templates, it enables efficient adaptation to diverse applications.

\subsection{Relationship between TAGs and KGs.}
Knowledge graphs (KGs) provide structured, ontology-driven data—where entities and relations adhere to well-defined schemas—and often function as powerful backbones for LLM applications such as factual retrieval and semantic question answering \cite{tan2024paths}. By associating minimal textual labels or short descriptions with symbolic triples ($i.e.$, “Entity–Relationship–Entity”), KGs facilitate entity disambiguation and precise relational reasoning in retrieval-augmented LLM workflows\cite{he2024g}. 
However, TAGs extend beyond these succinct entity–relation structures by attaching free-form or lightly structured text ($e.g.$, abstracts, user posts) to nodes, edges, or entire graphs. This richer textual context enables tasks that demand deeper language understanding, going beyond the typically fact-centric approach of KGs. For instance, TAGs excel in domains where entire documents or multi-sentence descriptions are crucial to graph-based learning, including document recommendation, social media analytics, and long-form knowledge exploration, illustrating how TAGs can fuse robust graph connectivity with extensive textual attributes.

%% file: s4-llm4tag.tex
\section{LLM for TAG} 
\label{sec:llm4tag}

\begin{figure}
    \centering
    \includegraphics[width=\linewidth]{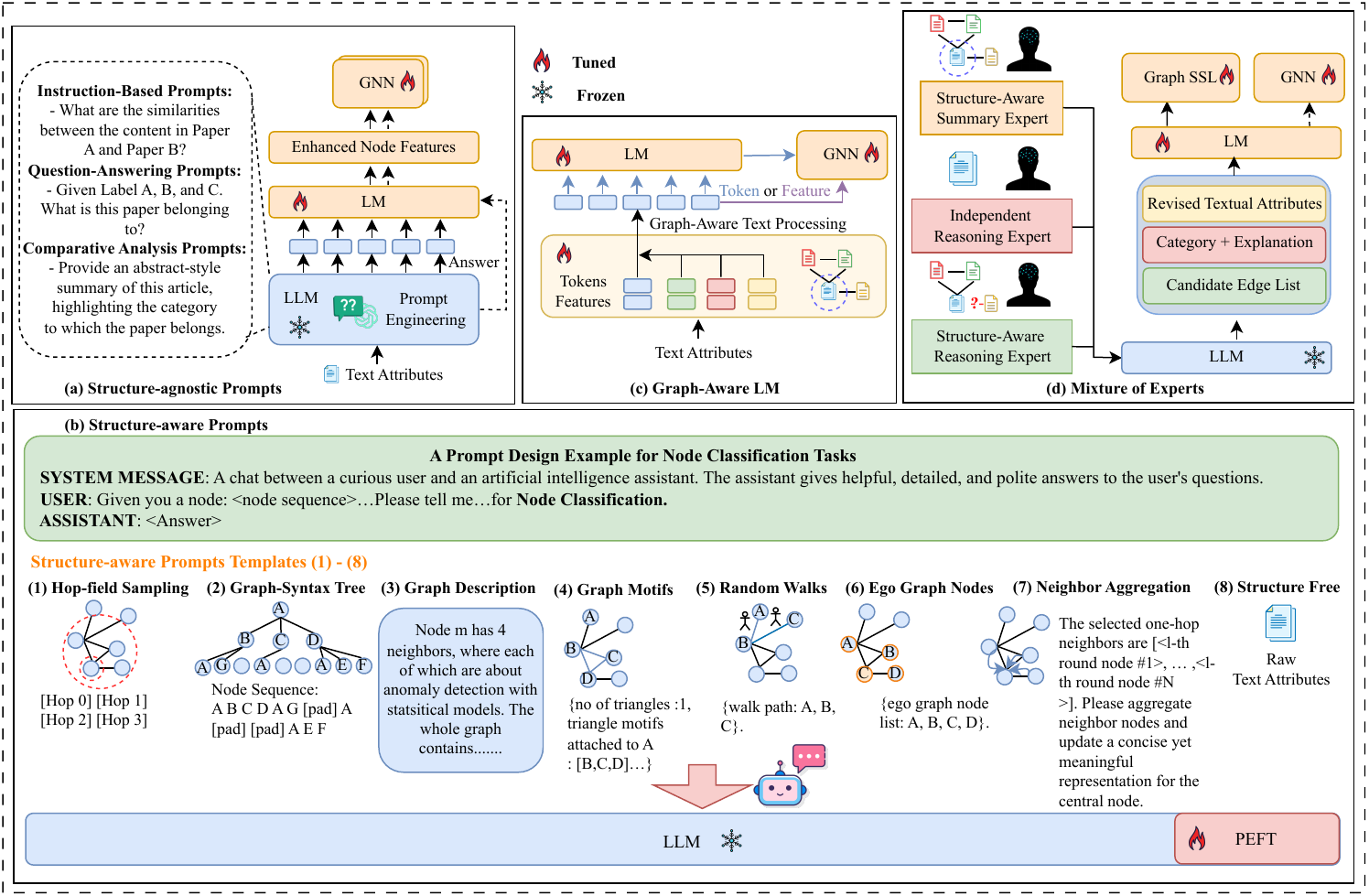}
    \caption{The illustration of sequential orchestration of LLM4TAG approaches, which contain (a) Structure-agnostic prompts, (b) Structure-aware prompts, (c) 
    Structure-aware language model (LM), and (d) Mixture of experts.}
    \label{fig:llm4tag-seq}
\end{figure}

In this section, we provide a comprehensive overview of how LLMs are orchestrated within the modeling of TAGs, emphasizing two complementary orchestration paradigms, \textbf{(1) sequential} and \textbf{(2) parallel orchestration} that enable LLMs to enhance TAGs' representation learning, semantic reasoning, and modality alignment. 
Furthermore, we also introduced \textbf{(3) TAG-specific pretraining models}, which leverage self-supervised pre-training, fine-tuning, and prompt-design techniques inspired by LLMs to translate language-model paradigms into the graph domain, thereby creating more expressive and generalizable TAG frameworks.
An intuitive overview of the proposed LLM4TAG orchestration pipelines are presented in \Cref{fig:llm4tag-seq}, \Cref{fig:parallel}, and \Cref{fig:tagpretrain}.

\subsection{Sequential Orchestration}
\label{subsec:3.1}
Sequential orchestration refers to strategies in which LLMs and TAG-based learning modules are applied in a stepwise manner, with one component enriching or transforming the input before the other, rather than being optimized jointly. Broadly, it can be categorized into two paradigms: one that applies LLMs as pre-processing or augmentation modules to enrich the textual attributes of TAGs prior to graph-based learning, and another that incorporates graph structural information directly into token design, enabling LLMs to perform topologically informed language reasoning.
More extensively, the sequential orchestration introduced in this section encompasses five representative approaches: 
\textbf{(1) Prompting on TAGs} asks the LLM task-specific questions while ignoring the underlying TAG topology.
\textbf{(2) Structure-agnostic prompts} rely solely on raw textual inputs without incorporating graph topology, treating each node or edge independently and enabling LLMs to generate enriched representations based purely on linguistic content.
\textbf{(3) Structure-aware prompts} incorporate structural graph signals, such as hop-based neighborhoods, motif patterns, or ego networks, into prompt templates, allowing LLMs to contextualize their outputs with localized topological information.
\textbf{(4) Structure-aware language models} embed graph structure directly into the encoding process of the LLMs, enabling joint modeling of graph topology and textual attributes within a unified representation space.
\textbf{(5) Mixture of experts} coordinates multiple specialized LLM modules, each responsible for distinct functions such as structural summarization, semantic reasoning, or context-aware inference, through a dynamic gating mechanism that activates the most relevant experts per instance.



\subsubsection{Prompting on TAGs}
\label{subsec:3.1.1}
Regarding TAGs, prompting can serve a dual purpose: it enables direct querying of task-specific questions to effectively leverage textual attributes, while also highlighting the critical integration of graph topological information.
In general, the informative text attributes generated by LLMs based on the given text prompts can be processed in two equally important ways. 
They can be directly utilized by downstream LLMs as predictors. 
Alternatively, as in the case of TAPE \cite{he2023harnessing}, the generated text attributes can be fine-tuned with task-specific SLMs, with the resulting enhanced node embeddings subsequently passed to GNNs for further processing and optimization.

Formally, given a TAG \( \mathcal{G} = (\mathcal{V}, \mathbf{A}, \bm{S}, \bm{T}) \) and a specific instruction prompt template \( \mathcal{T} \in \{\mathcal{T}(\cdot)\} \), we denote \( \bm{x} \) and \( \bm{y} \) as the LLM's input and target sentence, respectively. Then, prompt engineering can be formulated as:
\begin{align}
    P_\theta(\bm{y}_j \mid \bm{x}, \bm{y}_{<j}) &= \text{LLM}_\theta(\bm{x}, \bm{y}_{<j}), \quad \bm{x} = \text{Concatenate}(\mathcal{P}; \mathcal{I}; \mathcal{Q}), \label{eq:llm_pipeline} \\
    \mathcal{L}_\theta &= -\sum_{j=1}^{|y|} \log P_\theta(\bm{y}_j \mid \bm{x}, \bm{y}_{<j}), \label{eq:nll_loss}
\end{align}
Here, \(\mathcal{L}\) represents the negative log-likelihood (NLL) loss. 
The component \( \mathcal{I} \) of $\textbf{\textit{structure-aware prompts}}$ include the graph structure description derived from \( \mathcal{T}(\mathcal{V}, \mathbf{A}, \bm{S}, \bm{T}) \), leveraging various prompt templates to incorporate graph-specific information (More detail in \Cref{subsubstr-prompt}). 
In contrast, $\textbf{\textit{structure-agnostic prompts}}$ only use \( \mathcal{T}(\mathcal{V}, \bm{S}, \bm{T}) \). 
The task-specific instruction prefix \( \mathcal{P} \) and query \( \mathcal{Q} \) are designed to tailor the LLM for specific downstream tasks.

\subsubsection{Structure-agnostic Prompts.}
\hspace{10pt} 
\label{subsec:AgnosticPrompt}

Existing prompt formats, without explicitly incorporating the graph structure of input TAGs, often adopt an explanation-based enhancement framework, as demonstrated in TAPE \cite{he2023harnessing}. 
For example, in a node-level TAG representing academic citation networks, each node is characterized by the title and abstract of a paper. 
Taking node classification as an illustrative task, a commonly used instruction prefix \( \mathcal{P} \) and query \( \mathcal{Q} \) with \( \mathcal{I} \) in this context is:

\begin{enumerate}
    \item[\( \mathcal{I} \):] ``$\texttt{This is the node attributes [<Title, Abstract>]  of a literature.}$''

 \item[\( \mathcal{P} \):] ``$\texttt{Classify the node into one of these categories: [<All category>].}$''

 \item[\( \mathcal{Q} \):] ``$\texttt{{Which} arXiv CS sub-category does this paper
belong to? And provide your reasoning.}$''
\end{enumerate}

This type of prompt enables LLMs to generate outputs enriched with contextual information, where redundant content is filtered out, and task-relevant details are emphasized.
The enhanced embeddings derived from the LLMs or LMs can be further utilized with GNNs to improve TAGs training.
As highlighted by Chen et al. \cite{chen2024exploring}, incorrect predictions made by LLMs can sometimes appear reasonable within certain contexts due to their extensive pre-training on large-scale data, underscoring the critical role of explanations in understanding and justifying model outputs while potentially offering implicit insights.

{Due to its simplicity and effectiveness, this prompt format has been widely adopted in the design of numerous subsequent works \cite{liu2023one, huang2024can, wang2024large, chen2023label, ren2024representation}.}
As an example, LangGSL \cite{su2024bridging} designs prompts to mitigate noise in raw text data by summarizing input, incorporating symbolic elements such as emojis, and providing key factors as guidance, outperforming explanation-based prompts in some cases.
KEA \cite{chen2024exploring} prompts the LLMs to generate a list of knowledge entities along with their text descriptions and encodes them by fine-tuned LMs.
Shifting to edge-level tasks in TAGs, GraphEdit \cite{guo2024graphedit} directly reasons about potential dependencies between nodes using prompts.

For graph-level knowledge graphs (KGs) with text attributes on nodes and edges, LLM-SRR \cite{shi2024llm} uses prompt engineering to extract nuanced keywords and semantic features aligned with predefined targets, capturing critical user intent. Similarly, PROLINK \cite{wang2024llm} designs prompts incorporating task descriptions, entity types, and relation information to enable inductive reasoning across KGs without additional training.
Several studies \cite{wang2024can, han2023pive, guo2023gpt4graph, zhang2023llm4dyg, luo2023reasoning, huang2023can, zhao2023graphtext, fatemi2023talk, ye2023natural} find that LLMs exhibit preliminary capacity for graph reasoning, particularly under simpler tasks such as cycle detection or shortest path. 
These works suggest that, even without extensive architectural modifications, LLMs can recognize and reason about graph-structured data when properly prompted.

\subsubsection{Structure-aware Prompts}
\hspace{10pt}
\label{subsubstr-prompt}


\noindent
Structure-aware prompts explicitly embed graph topology, node-edge attributes, and subgraph relationships into the prompt, providing richer context compared to structure-agnostic prompts that rely solely on textual attributes (\Cref{fig:llm4tag-seq}a). 
To effectively translate graph topological information into tokens interpretable by downstream LLMs, existing works design structure-aware prompts based on two key considerations.
First, templates \( \mathcal{I} \) are crafted to standardize the representation of graph topology. 
As shown in \Cref{fig:llm4tag-seq}b, existing graph description templates \( \mathcal{I} \) can be categorized into eight distinct types, each designed to address specific task requirements effectively.
Secondly, by integrating these templates with the task-specific instruction prefix \( \mathcal{P} \) and query \( \mathcal{Q} \), LLMs can be effectively utilized as predictors for downstream tasks.

For example, in node classification on a TAG of academic citation networks, a commonly used instruction prefix \( \mathcal{P} \) and query \( \mathcal{Q} \) with \( \mathcal{I} \) based on hop-field template (which can follow any of the templates (1) to (7) below based on need) is:

\begin{enumerate}
    \item[$\mathcal{I}$:]  \sloppy ``\texttt{Node\_1 [<Title\_1, Abstract\_1>] is connected with Node\_4 [<Title\_4, Abstract\_4>], and Node\_7 [<Title\_7, Abstract\_7>] within one hop.}''

\item[\( \mathcal{P} \):] \sloppy
``\texttt{Classify the node into one of these categories: [<All category>], considering the  link relationships between the nodes. }``

\item[\( \mathcal{Q} \):] ``$\texttt{{Which} arXiv CS sub-category does this paper
belong to? And provide your reasoning.}$''
\end{enumerate}


\begin{enumerate}
    \item \textbf{Hop-field template.}
Templates based on hop-field information are designed to encode the structural context of a graph by incorporating relationships within a specified number of hops around a central node.


Similar ideas have also been explored in MuseGraph \cite{tan2024musegraph}, GNN-RAG \cite{mavromatis2024gnn}, and LangTopo \cite{guan2024langtopo}, leveraging hop-field templates to incorporate graph structural information into task-specific prompt design.
It should be noted that utilizing up to 3-hop connectivity is sufficient for excellent performance \cite{kipf2016semi, velivckovic2017graph, hamilton2017inductive} while information beyond 3-hop typically owns a minor impact on improvement and might even lead
to negative effects \cite{ chen2020measuring, rong2019dropedge, su2024simple}.

\item \textbf{Graph-syntax tree template.} 
The graph-syntax tree template bridges relational and sequential data by organizing nodes, their textual attributes, and inter-node relationships into a hierarchical tree structure. 
Starting from a root node, child nodes are sequentially connected based on their relationships, with edges labeled by relational attributes. 
LLaGA \cite{chen2024llaga}, HiCom \cite{zhang2024hierarchical}, and GRAPHTEXT \cite{zhao2023graphtext} leverage graph-syntax trees to represent complex graph data. By traversing these trees, natural language sentences are generated and fed into LLMs, enabling effective graph reasoning and representation. 
\item \textbf{Graph-text pair template.} Templates for {{graph-text pair}} are designed to incorporate structural information, capturing both node-level details and the overall graph structure information {\cite{wei2024gita, lin2024unleashing, zhao2023gimlet}}. 
By summarizing node attributes, connectivity, and global properties such as topology and relational patterns, these templates provide a comprehensive representation, making them particularly effective for complex domains like molecular studies \cite{shi2023relm, liu2023molca, liu2024git, cao2023instructmol, xu2024llm}.
\item \textbf{\textbf{Graph motifs template.}} 
GRAPHTMI \cite{das2023modality} and PROLINK \cite{wang2024llm} utilize graph motifs templates to encode the presence and relationships of specific motifs, such as triangles, cycles, or cliques, within TAGs. 
The core idea of graph motifs is to identify recurring subgraph patterns that represent meaningful structural components in the graph. By capturing these patterns, motifs provide a compact yet expressive representation of graph topology, which can be enriched with textual attributes to enhance downstream tasks.


\item \textbf{Random walk-based template.} MuseGraph \cite{tan2024musegraph}
and GraphCLIP \cite{zhu2024graphclip} utilize 
{random walk-based} templates to capture graph structure by summarizing the connectivity patterns and local neighborhood relationships of a graph through random walk sequences.
\item {\textbf{Ego-based template.}}  Several works \cite{tan2024musegraph, shi2024llm, tang2024graphgpt} construct 
$h$-hop subgraphs around each central node using random neighbor sampling, effectively encoding local structural information into inputs for LLMs.
PromptGFM \cite{gnngraph} employs one-hop neighbor sampling to represent graph structure and uses straightforward prompts to simulate a flexible 
\item {\textbf{Neighbor aggregation template.}} 
The mechanism aims to achieve message passing entirely through textual descriptions rather than traditional GNN frameworks or explicit neighbor embeddings based Hop-Field Overview Template used in LLaGA.
\end{enumerate}
\noindent



Compared to structure-agnostic prompts, structure-aware prompts are particularly advantageous for addressing graph-specific tasks, such as planning in robotics \cite{huang2022language, andreas2022language}, multi-hop question
answering or knowledge probing \cite{adhikari2020learning, ammanabrolu2021learning}, structured commonsense reasoning \cite{tandon2019wiqa, madaan2022language}, and more.
Based on the designed benchmarks, incorporating structural information, such as neighborhood summarization, has been shown to enhance GPT performance on node-level tasks, with studies reporting slight gains \cite{chen2024exploring, hu2023beyond} and significant improvements, including a 10\% accuracy increase on ogbn-arxiv with 2-hop summarization \cite{guo2023gpt4graph}.
Studies like \cite{huang2024can} and \cite{guo2023gpt4graph} demonstrate that structural prompts, including neighborhood homophily encoding and role-based designs, allow LLMs to process graph structure as linearized text while achieving competitive performance on structural reasoning tasks. Techniques like natural language prompts encoding multi-hop connections \cite{ye2023natural} and adjacency-based chain-of-thought prompting \cite{fatemi2023talk} further validate the ability of LLMs to bridge the gap between textual and structural representations.

\subsubsection{Structure-aware Language Models}
\hspace{10pt}
\label{subsub:structureLM}

\noindent
As illustrated in \Cref{fig:llm4tag-seq}c, the methods discussed in this section try to integrate graph topology information into the representation learning phase of language models ($e.g.$, Transformers such as BERT).
These methods address the limitations of traditional feature engineering approaches, where numerical node features extracted from raw data remain graph-agnostic, preventing the full utilization of the correlations between graph topology and node attributes.



Taking a node-level TAG \(\mathcal{G} = (\mathcal{V}, \mathcal{E}, \bm{S})\) as an example, the representation learning process leverages both the adjacency matrix \(\bm{A}\) and the node-level textual attributes \(\bm{S}\) to generate enhanced node embeddings \(\bm{H} \in \mathbb{R}^{n \times d'}\), formulated as:  
\begin{equation}
    \bm{H} = f_{\text{Transformer}}(\bm{A}, \bm{S}; \bm{\theta}),
\end{equation}
where \(f_{\text{Transformer}}\) represents a Transformer model adapted to incorporate both graph topology and textual attributes, and \(\bm{\theta}\) denotes its learnable parameters.


GIANT \cite{chien2021node} as a pioneering framework, leverages a self-supervised task called neighborhood prediction, which models graph structure as a multi-label classification problem. By fine-tuning a language model like BERT with this graph-structured supervision, GIANT generates node embeddings that seamlessly integrate raw text attributes with graph topology, enabling enhanced representation learning.

Furthermore, several studies have proposed different mechanisms to better integrate graph structure into the learning process. 
For instance, Edgeformers \cite{jin2023edgeformers} integrate network information into each Transformer layer while encoding textual edges, and subsequently aggregate these contextualized edge representations within each node’s ego-graph. 
This design facilitates more effective learning of the central node’s representation for downstream tasks like node classification and link prediction.
Similarly, GraphBridge \cite{wang2024bridging} unifies local and global topological perspectives by leveraging contextual textual information. It selectively retains crucial tokens based on both graph structure and task-specific relevance, ultimately refining node representations. 
In contrast to approaches that jointly optimize structure and text, GRAD \cite{mavromatis2023train} employs a shared language model for bidirectional optimization. Specifically, it utilizes a GNN teacher model to encode both graph topology and node attributes, producing graph-informed soft labels. These labels then guide a graph-free student model ($e.g.$, BERT) through a distillation process, enabling the student to internalize structural correlations and global graph context within its textual embeddings for downstream tasks.
LLM-BP \cite{wang2025llmbp} turns a text-attributed graph into a “mini-MRF” in which task-aware node embeddings from an LLM serve as unary potentials and an LLM-estimated homophily constant sets the pairwise potentials; a few belief-propagation steps then yield zero-shot labels that beat prior TAG baselines without any gradient training.
GraphEval \cite{feng2025grapheval} leverages the topological structure of a viewpoint graph that decomposes research ideas into fine-grained viewpoint nodes linked by LLM derived edges and propagates quality signals via label propagation or a lightweight GNN. 
Dr.E \cite{liu2025multi} connects graph data to LLMs through a chain of multi view structural enhancement, dual residual vector quantization, and token level alignment, turning graph structure into natural language tokens that yield interpretable, efficient, and robust gains on tasks such as node classification.


\subsubsection{Mixture of Experts}
\hspace{10pt}
\label{subsub:MoE}

\noindent
As illustrated in \Cref{fig:llm4tag-seq}d, in the context of TAGs, mixture of 
experts (MoE) models \cite{rajbhandari2022deepspeed, du2022glam} analyze node and edge attributes by leveraging a set of specialized expert modules, each focusing on distinct aspects of the graph data. 
These experts dynamically incorporate graph topology and textual semantics to generate task-specific insights. 
Through a gating mechanism, MoE models activate the most relevant experts for a given input, enabling efficient and context-aware operations such as adaptive node classification, contextual graph exploration, and automated reasoning.

GAugLLM \cite{fang2024gaugllm} introduces a MoE framework with three specialized experts: ``Structure-Aware Summarization'', which captures local structural context to summarize node attributes; ``Independent Reasoning'', which focuses on high-level semantic predictions through open-ended prompts; and ``Structure-Aware Reasoning'', which integrates neighbor relationships and graph connections into reasoning prompts. 
This framework effectively enhances self-supervised graph learning by aligning textual and structural information for robust node representations.

Existing works on MoE primarily focus on text-attributed knowledge graphs, offering diverse approaches to balance exploration and exploitation, as well as reasoning and decision-making tasks. WESE \cite{huang2024wese} balances exploration and exploitation by employing a cost-effective weak agent for knowledge acquisition, which is stored in a graph-based structure to guide efficient exploitation. Similarly, LociGraph \cite{cho2024locigraph} enables autonomous extraction of structured information from non-public web environments, emphasizing the effective use of non-traditional data sources. However, Wu et al. \cite{wucan} takes a different direction by integrating GNNs with LLMs to address decision-making challenges in large task graphs, demonstrating superior performance with minimal training through efficient alignment of graph and text representations.
In contrast, GA \cite{wang2023graph} focuses on graph reasoning by combining symbolic reasoning and textual transformations to deliver interpretable predictions.


\subsubsection{Discussion}
\label{subsec31:discussion}
\hspace{10pt}

\noindent
The five representative approaches of sequential orchestration share a unifying principle: they all treat LLMs as modular enhancers that can be strategically positioned before or alongside TAG-based learners to enrich representation learning. A common thread is the reliance on natural language prompts or token-level adaptations as the interface between textual attributes and graph structure. In this way, they leverage the pretrained linguistic knowledge of LLMs to augment TAG modeling with richer semantics and, in some cases, structural awareness.

Despite this commonality, the approaches differ in their level of structural integration and design philosophy. Prompting on TAGs and structure-agnostic prompts emphasize simplicity and broad applicability, focusing primarily on textual enrichment while leaving graph reasoning to downstream models. Structure-aware prompts and structure-aware language models progressively embed more topological signals, either through handcrafted prompt templates or through architectural adaptations that encode graph connectivity directly into the LLM. Mixture-of-experts models stand apart in their modularity, coordinating multiple specialized LLMs to balance semantic reasoning and structural summarization, offering a more flexible but complex orchestration paradigm.

From a computational perspective, these methods span a wide spectrum of costs. Prompting on TAGs and structure-agnostic prompts are lightweight and often require minimal additional computation beyond standard LLM inference, making them attractive for scalability. 
Structure-aware prompts introduce overhead proportional to the complexity of topology encoding (e.g., hop-based neighborhoods or motif descriptions), while structure-aware language models incur higher costs due to architectural modifications and fine-tuning requirements. Mixture-of-experts methods, although potentially more parameter-efficient via sparse activation, introduce coordination and routing overhead that can increase both memory and latency. 
Overall, these approaches reveal a trade-off between structural expressiveness and computational efficiency, underscoring the need for adaptive orchestration strategies that balance accuracy and scalability across diverse TAG applications.

\begin{wrapfigure}{r}{11cm}
\centering
\includegraphics[width=0.67\textwidth]{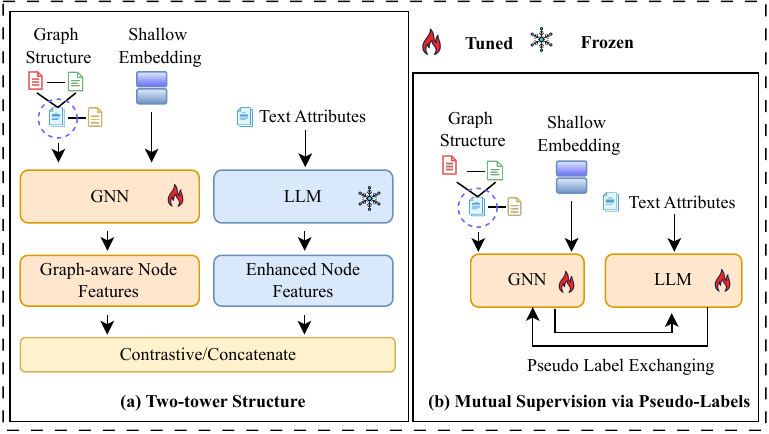}
\caption{The illustration of parallel orchestration of LLM4TAG approaches, which contain (a) Two-tower structure, and (b) Mutual supervision via pseudo-labels.}
\label{fig:parallel}
\end{wrapfigure}

\subsection{Parallel Orchestration}
\label{subsec:parallel}

Parallel orchestration refers to frameworks where LLMs and TAGs operate as separate yet collaborative modules, each specializing in processing a different modality: text attributes and graph structure, respectively. 
Unlike sequential orchestration, which feeds one modality's output into the other, parallel orchestration maintains architectural independence while enabling interaction through shared objectives or joint embedding spaces.
Specifically, as illustrated in \Cref{fig:parallel}, we identify two major categories of parallel orchestration: 
\textbf{(1) Two-tower structure} independently encodes text and graph inputs of TAGs using LLMs and GNNs, then aligns their embeddings in a shared latent space. 
This decoupled design enables flexible cross-modal learning, where contrastive objectives are often used to match graph structures with their textual descriptions.
\textbf{(2) Mutual supervision via pseudo-labels} enables iterative interaction between LLMs and GNNs by exchanging pseudo-labels. 
Each model refines the other’s predictions, treating text and graph as complementary sources of supervision.

\subsubsection{Two-tower Structure}
\hspace{10pt}
\label{subsub:2tower}

\noindent
Two-tower models \cite{su2023beyond, li2023blip, karpukhin2020dense, feng2021zero} offer a scalable and efficient architecture. 
By independently encoding inputs from two domains ($i.e.$, graphs, and text) into a shared embedding space, they enable effective similarity computations using metrics like dot product or cosine similarity. 
Following the two-tower architecture, recent works on TAGs leverage LLMs and GNNs as two independent modules (\Cref{fig:parallel}a).

In general, LLMs process textual information, while GNNs encode graph structures, these embeddings are subsequently fused in the shared space to produce enriched graph representations.
SIMTEG \cite{duan2023simteg} introduces a decoupled framework for textual graph learning, separating textual embedding generation from graph structure learning. 
It utilizes pre-trained language models for extracting high-quality text representations and GNNs to incorporate graph structural information. 
The two models are guided with consistent loss
function, $e.g.$, link prediction, or node classification.
\cite{selvam2024can} advanced node embedding quality by leveraging prompt engineering and sharing the same architecture with SIMTEG.
LATEX-GCL \cite{yang2024latex} employed graph self-supervised learning by contrasting graphs with shallow node embeddings against graphs with enhanced node embeddings in the latent space.

GraphCLIP \cite{zhu2024graphclip}, GNP \cite{tian2024graph}, and G2P2 \cite{wen2024prompt} extend the foundational principles of CLIP \cite{radford2021learning} by treating graphs and their associated textual descriptions as distinct modalities. To align the latent spaces of graph representations and text embeddings, a contrastive loss is employed:
\begin{scriptsize}
    \begin{equation}
\mathcal{L}_{\text{contrast}} = - \log \frac{\exp(\cos(h_G, h_{d_G}) / \tau)}{\exp(\cos(h_G, h_{d_G}) / \tau) + \exp(\cos(h_G, h_{\tilde{d}_G}) / \tau)},    
\end{equation}
\end{scriptsize}
where \(h_\mathcal{G}\) and \(h_{d_\mathcal{G}}\) are the latent representations of graph \(\mathcal{G}\) and its paired description \(d_\mathcal{G}\), \(\tilde{d}_\mathcal{G}\) is a negative sample (a description not paired with \(\mathcal{G}\)), and \(\tau\) is the temperature hyperparameter.
These methods employ contrastive pre-training techniques to align graph substructures with their textual summaries in a shared latent space. 
This alignment not only bridges the modality gap but also enhances the cross-domain adaptability of graph representations, enabling robust zero-shot and few-shot transferability across diverse TAG datasets.
As a widely researched area, in the study of molecular graphs, molecules inherently have a natural graph structure that can be effectively learned using models like GIN \cite{xu2018powerful}. 
paired with molecular graphs, we often have abundant textual annotations, such as chemical descriptions, SMILES (Simplified Molecular Input Line Entry System) strings~\cite{weininger1988smiles}, or functional property labels.
Methods such as Text2Mol \cite{edwards2021text2mol}, MoMu \cite{su2022molecular}, MoleculeSTM \cite{liu2023multi}, and ConGraT \cite{brannon2023congrat} focus on leveraging this multimodal information, demonstrating significant advancements in integrating textual and structural features for molecular representation study by leveraging contrastive learning.

\subsubsection{Mutual Supervision via Pseudo-Labels}
\hspace{10pt}
\label{subsub:mutualPseduoL}

\noindent
Unlike two-tower architecture methods, the iterative alignment approach, illustrated in \Cref{fig:parallel}b, treats both modalities symmetrically while introducing a distinct training paradigm.
GLEM \cite{zhao2022learning} leverages an Expectation-Maximization framework with two encoders: an LM uses local textual information to model label distributions, while a GNN encoder uses labels and text from surrounding nodes to capture global conditional label distributions. 
The two encoders iteratively refine their representation spaces by generating pseudo-labels for each other. 
However, GLEM is primarily built on the assumption that the provided graph structure is complete and noise-free, which is difficult to guarantee in real-world applications. 
A gap addressed by LangGSL \cite{su2024bridging} and LangTopo \cite{guan2024langtopo} integrates graph structure learning to enhance representation quality, while extending applicability to challenging scenarios such as adversarial attacks or completely absent graphs.
What's more, GraphEval \cite{feng2025grapheval} is a lightweight graph–LLM framework that first decomposes each research idea into fine-grained “viewpoint” nodes via a small prompted LLM and then links them into a viewpoint-graph using embedding-based similarity, enabling both local semantic scoring and global relational reasoning.

\subsubsection{Discussion}
\hspace{10pt}

\noindent
Both families aim to align textual semantics with graph structure in a shared representation space, while preserving the unique strengths of each modality: LLMs contribute rich linguistic priors, whereas GNNs capture inductive structural biases.
Two-tower methods emphasize decoupled encoding with a global contrastive or task-specific loss. This design is simple, scalable, and efficient for deployment. In contrast, iterative alignment introduces bidirectional supervision through pseudo-labels, which can enhance cross-modal consistency and improve label efficiency, particularly when coupled with graph structure learning.

Two-tower training scales linearly with corpus size, benefits from batched contrastive objectives, and is typically latency-friendly at inference (single forward pass per tower). Iterative alignment, however, requires repeated pseudo-labeling and re-training cycles. When augmented with structure learning, it achieves robustness to noisy or incomplete graphs, but at the expense of additional computation.

\subsection{Pre-trained Model for TAGs}
\label{subsec:pretrainTAG}

In this section, we provide a comprehensive overview of techniques for TAG-based pre-training models, which aim to unify the representation of text attributes and graph structures within a single framework.
These models are designed to capture rich semantic and structural information during pre-training, enabling effective transfer to downstream tasks through fine-tuning, graph prompting, or other adaptation strategies. 
The discussion is organized into four main categories as illustrated in \Cref{fig:tagpretrain}: \textbf{(1) TAG Transformers, (2) Graph foundational models, (3) Fine-tuning, and (4) Graph prompt tuning.}
For each category, we further delineate subcategories based on shared methodologies and underlying principles, offering a structured analysis of the state-of-the-art approaches.


\subsubsection{{TAG Transfromers}}
\label{subsec:graphTransformer}
\hspace{10pt}

\begin{figure}
    \centering
    \includegraphics[width=\linewidth]{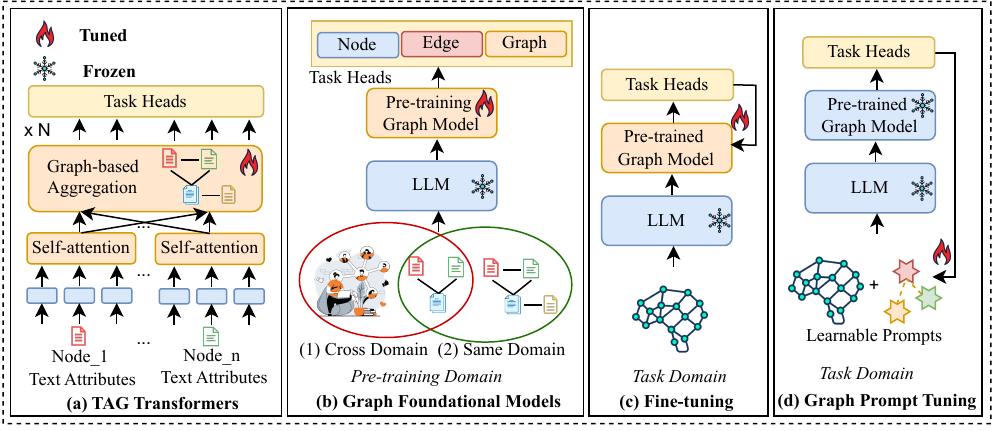}
    \caption{The illustration of pre-trained models for TAGs of LLM4TAG approaches, which contain (a) TAG Transformers, (b) Graph foundation models, (c) Fine-tuning, (d) Graph prompt tuning.}
    \label{fig:tagpretrain}
\end{figure}

\noindent
As illustrated in \Cref{fig:tagpretrain}a, TAGs Transformers adapt the Transformer \cite{vaswani2017attention} framework to graph-structured data by using self-attention mechanisms that capture both text attributes and graph structures. 
In self-attention, the attention score between nodes \( v_i \) and \( v_j \) is computed as \( a_{ij} = \frac{\mathbf{Q}_i^\top \mathbf{K}_j}{\sqrt{d}} \), where \( \mathbf{Q}_i \) and \( \mathbf{K}_j \) are the query and key vectors for nodes \( v_i \) and \( v_j \), and \( d \) is the feature dimension. The attention weight \( \alpha_{ij} \) is then normalized as \( \alpha_{ij} = \frac{\exp(a_{ij})}{\sum_{k \in \mathcal{N}(v_i)} \exp(a_{ik})} \), where \( \mathcal{N}(v_i) \) denotes the neighbors of node \( v_i \). 
Node representations are updated by aggregating information from neighboring nodes, given by \( \mathbf{h}_i = \sum_{j \in \mathcal{N}(v_i)} \alpha_{ij} \mathbf{V}_j \), where \( \mathbf{V}_j \) represents the feature vector of neighbor \( v_j \). 
Multi-head attention is employed to capture diverse relationships by performing multiple attention operations in parallel: \( \mathbf{H}_i = \text{Concat}(\mathbf{h}_i^{(1)}, \dots, \mathbf{h}_i^{(h)}) \), where \( h \) represents the number of attention heads. 


Based on the basic idea of Transformers, the PATTON \cite{jin2023patton} framework introduces two pre-training strategies: network-contextualized masked language modeling and masked node prediction, both designed to capture the inherent dependencies between textual attributes and the network structure. Similarly, the TextGT \cite{yin2024textgt} framework for aspect-based sentiment analysis processes TAGs using a double-view approach, where GNNs model word relationships in the graph view, and Transformer layers capture the sequential structure of the text. Additionally, it introduces TextGINConv, a specialized graph convolution that incorporates edge features for more expressive node representations, thereby enhancing the integration of structural and textual information.

In contrast, the GSPT \cite{song2024pure} framework treats graph structure as a prior and leverages the unified feature space of LLMs to learn refined interaction patterns that generalize across graphs. It samples node contexts through random walks and applies masked feature reconstruction using a standard Transformer to capture pairwise proximity in the LLM-unified feature space, while ENGINE \cite{zhu2024efficient} offers a parameter- and memory-efficient fine-tuning method.
Some models share similar ideas but differ across application domains. For instance, GIMLET \cite{zhao2023gimlet} is designed for molecule property prediction. Meanwhile, GraphFormers \cite{yang2021graphformers} employs a nested structure, and JointGT \cite{ke2021jointgt} and TGformer \cite{shi2024tgformer} focus on knowledge graphs, whereas TG-Transformer \cite{zhang2020text} is tailored for document classification. These advancements highlight the diverse applications of graph-transformer-based frameworks in integrating structural and textual information.
Recently, GraphGPT2 \cite{zhao2025graphgpt} presents the Graph Eulerian Transformer which follows an Eulerian path through a graph, writes the visited nodes, edges, and attributes as a reversible token sequence, and feeds this sequence into an ordinary Transformer in the same way that text is processed. Using next-token and scheduled masked-token prediction during generative pre-training, the model can be grown to billions of parameters and, after simple fine-tuning, delivers strong results on graph, edge, and node prediction tasks






\subsubsection{Graph Foundational Models}
\label{subsec:SSL}
\hspace{10pt}

\noindent
Foundation models leverage large-scale deep learning and transfer learning to achieve emergent capabilities and strong performance across a wide variety of tasks.  
In the TAG domain, as illustrated in \Cref{fig:tagpretrain}b, graph foundation models (GFMs) are pre-trained on vast and diverse collections of multi-graph data drawn from domains that can be consistent or different with each other, learning to encode rich structural and attribute information.  
By employing cross-domain pre-training, these models produce transferable representations at the node, edge, and whole-graph levels, which can be fine-tuned or adapted to a broad spectrum of downstream graph tasks.

In this section, we introduce GFMs which are pre-trained via self-supervised learning (SSL) with objectives specifically tailored to TAGs.
Specifically, in TAG-based SSL, a GNN \(\psi(\cdot)\) is trained on task-agnostic objectives, such as contrastive learning or predictive tasks, to learn generalizable graph representations. This process can be expressed as:
\begin{equation}
\psi^* = \arg\min_\psi \mathbb{E}_{\mathcal{G} \sim \mathcal{D}} \mathcal{L}_{\text{SSL}}(\psi(\mathcal{G})), 
\end{equation}
where \(\mathcal{L}_{\text{SSL}}\) denotes the self-supervised loss, \(\mathcal{G}\) is a graph sampled from the TAGs dataset \(\mathcal{D}\), and \(\psi^*\) represents the pre-trained model.

Recent works collectively lay the foundations for general‐purpose graph models by advancing various complementary themes. 
OFA \cite{liu2023one} unifies diverse graph tasks—node, link, and whole-graph classification—into a single text-attributed representation, enabling in-context learning via a Nodes-of-Interest prompt without any fine-tuning. 
UniGLM \cite{fang2024uniglm} scales contrastive pretraining across multiple TAGs, introducing adaptive sampling and a lazy update mechanism to align graph structure with textual embeddings in a frozen language encoder. 
GraphCLIP \cite{zhu2024graphclip} leverages LLM-generated natural-language summaries of subgraphs and an invariant contrastive objective to achieve robust zero-shot transfer across domains. 
Adapter-based methods (LLaGA \cite{chen2024llaga}, GraphAdapter \cite{li2024graph}, TEA-GLM \cite{wang2024llms}) then demonstrate how lightweight GNN modules or linear projectors can bridge graph representations and LLM token spaces, supporting both zero- and few-shot inference. What's more, PATTON \cite{jin2023patton} enriches masked-language-model pre-training with explicit graph structure via masked node prediction, and Text-Space GFMs and GLIP-OOD \cite{xu2025glip} provide standardized benchmarks and zero-shot out-of-distribution detection, respectively. 
By unifying task formats, exploiting contrastive and adversarial objectives, designing prompt- and token-based conditioning, and embedding structural cues into core pretraining, these contributions jointly chart a research roadmap toward versatile, high-quality graph foundation models.

GraphMaster \cite{du2025graphmaster} extends this roadmap by tackling the foundational challenge of scarce graph corpora: it orchestrates four specialized LLM “agents” (Manager, Perception, Enhancement, Evaluation) to iteratively synthesize semantically rich, TAGs under tight data constraints, combining structural optimization with natural-language attribute generation.
AutoGFM \cite{chen2025autogfm} is an automated graph foundation model that learns a graph-to-architecture mapping inside a weight-sharing super-network, tailoring the GNN backbone to each specific domain and task. 
It fuses a disentangled contrastive graph encoder, invariant-guided architecture customization, and a curriculum-based search strategy on diverse node-, link-, and graph-level benchmarks.
In contrast, MDGFM \cite{wang2025mdgfm} aligns both node features and graph topologies from diverse source domains—via shared/domain tokens, an adaptive balance token, and graph-structure learning — to learn domain-invariant representations.
This unified pre-training plus dual-prompt tuning framework enables robust, few-shot knowledge transfer to unseen homophilic and heterophilic graphs, outperforming prior multi-domain GNN baselines and remaining resilient to noise and adversarial attacks.

\subsubsection{Fine-tuning}
\hspace{10pt}
\label{subsec:SSLft}

\noindent
Fine-tuning is commonly employed to adapt pre-trained GFMs to downstream tasks by updating model parameters using labeled data.
Specifically, it involves adapting the model \(\psi^*(\cdot)\) to a specific downstream task \(\mathcal{T}\) using labeled data \(\mathcal{D_T}\), by optimizing the following objective:
\begin{equation}
\psi_{\mathcal{T}}^* = \arg\min_\psi \mathbb{E}_{\mathcal{G}, y \sim \mathcal{D_T}} \mathcal{L}_{\mathcal{T}}(\psi(\mathcal{G}), y),    
\end{equation}
where \(\mathcal{L}_{\mathcal{T}}\) is the task-specific loss, and \(y\) represents the task labels. 
In TAG-based GFMs, both the structural information and raw text attributes should be considered for effective representation learning.
For methods that leverage fine-tuning with additional labeled data, HASH-CODE \cite{zhang2024high} takes a novel approach by integrating GNNs and LLMs into a unified model, applying high-frequency-aware contrastive learning to ensure more distinctive embeddings. Similarly, NRUP \cite{kuang2023unleashing} proposes a node representation update pre-training architecture based on co-modeling text and graph, where hierarchical graph construction and self-supervised tasks contribute to improved node feature updates and enhanced model generalization.

In contrast, Grenade \cite{li2023grenade} introduces a graph-centric language model that focuses on optimizing graph-centric contrastive learning and knowledge alignment, demonstrating superior performance in capturing both textual semantics and structural information. Meanwhile, GAugLLM \cite{fang2024gaugllm} and LATEX-GCL \cite{yang2024latex} leverage LLMs to augment textual features, addressing key challenges such as information loss, semantic distortion, and the misalignment between text and graph structures, thus improving pre-training performance for downstream tasks.

GraphLoRA \cite{10.1145/3690624.3709186} is a parameter-efficient fine-tuning framework that adapts Low-Rank Adaptation (LoRA) \cite{hu2022lora} to graph neural networks (GNNs). By injecting low-rank trainable matrices into select layers of pre-trained GNNs, GraphLoRA enables task-specific adaptation with minimal computational overhead and reduced memory footprint. This design allows for rapid deployment across multiple downstream tasks without retraining the full model, making it particularly suitable for scenarios with limited resources or requiring multi-task support.








\subsubsection{Graph Prompt Tuning}
\hspace{10pt}
\label{subsec:SSLprompt}

\noindent
Building on the concept of soft prompts \cite{brown2020language} in language models, where learnable embeddings guide pre-trained models toward tasks without altering parameters, {graph prompting \cite{sun2023all, liu2023graphprompt, fang2024universal, sun2022gppt, fu2025edge, 10.1145/3690624.3709219}} introduces a novel paradigm, termed \textbf{"pre-training, graph prompting, and predicting"}, facilitating seamless alignment between pre-trained GNNs and downstream tasks.
Given a frozen pre-trained GNN \(\psi^*(\cdot)\), trained through self-supervised learning (SSL) to capture task-agnostic representations of graph data, a prompting function \(g(\cdot)\) modifies an input graph \(\mathcal{G}\) into an optimized form \(g(\mathcal{G})\). The prompt module \(\mathcal{M}(\mathcal{G}, \mathcal{G_P})\) transforms input graphs into task-specific representations that align with pre-training objectives. This module $\mathcal{M}$ incorporates the original graph \(\mathcal{G}\), a prompted graph \(\mathcal{G_P}\), and learnable parameters ($e.g.$, modified adjacency or feature matrices) to ensure alignment. The relationship can be formalized as:
\begin{equation}
    \psi^*\left( \mathcal{M}(\mathcal{G}, \mathcal{G_P}) \right) = \psi^*\left( g(\mathcal{G}) \right) + O_{\mathcal{P}\psi},
    \label{eq:graph_prompting}
\end{equation}
where \(O_{\mathcal{P}\psi}\) represents the error bound between the representations of the prompted and optimally transformed graphs \cite{zi2024prog}.
Compared to fine-tuning, graph prompting is more suitable for few-shot scenarios and is more efficient as it avoids modifying the parameters of the pre-trained model. Additionally, existing graph prompting frameworks enable it to simultaneously handle tasks at various levels of graph representation.


OFA \cite{liu2023one} presents the first general framework for unifying diverse graph data by representing nodes and edges with natural language descriptions. It leverages language models to encode cross-domain text attributes of TAGs into a shared embedding space, facilitating consistent feature representation. Furthermore, OFA introduces a novel graph prompting paradigm, where task-specific substructures are appended to input graphs, enabling the framework to address various tasks without requiring fine-tuning.

Building on similar ideas, recent advancements have explored graph prompting to enhance few-shot and zero-shot learning on TAGs. For instance, Hound \cite{wang2024hound} introduces novel augmentation techniques, such as node perturbation and semantic negation, to provide additional supervision signals and improve zero-shot node classification. In contrast, ZeroG \cite{li2024zerog} focuses on cross-dataset zero-shot transferability, employing prompt-based subgraph sampling and lightweight fine-tuning to tackle challenges like feature misalignment and negative transfer.
On the other hand, G-Prompt \cite{huang2023prompt} combines a learnable graph adapter with task-specific prompts to seamlessly integrate textual and structural information, delivering enhanced interpretability and performance for few-shot learning. Similarly, P2TAG \cite{zhao2024pre} incorporates self-supervised learning through masked language modeling and graph pre-training, achieving significant accuracy improvements over existing TAG methods with graph prompt tuning.
EdgePrompt \cite{fu2025edge} adapts a frozen, pre-trained GNN to new tasks by learning small, trainable vectors on each graph edge—aggregated during message passing—to inject task-specific signals without altering the backbone weights.

While graph prompt tuning provides an efficient approach for adapting pre-trained GFMs to downstream tasks with minimal parameter updates, one limitation of graph prompt tuning is its lack of inherent support for zero-shot learning, unless it is extended with techniques such as virtual class construction, where the embeddings of textual descriptions of target categories are used to guide the model, as demonstrated in ZeroG.
Moreover, when the number of shots increases, prompt tuning often struggles to match the performance of full fine-tuning. 
These challenges highlight the need for further research to improve the adaptability and scalability of prompt-based methods, especially in the context of GFMs.

\subsubsection{Discussion}
\hspace{10pt}

\noindent
Pre-trained models for TAGs share the overarching goal of learning universal and transferable representations that unify textual semantics with graph topology, yet they pursue this objective through distinct strategies with varying trade-offs. TAG Transformers extend self-attention directly to graph structures, capturing both local and global dependencies but often incurring substantial training costs and scalability challenges. Graph foundation models (GFMs), in contrast, rely on large-scale self-supervised pre-training across diverse corpora, offering strong zero- and few-shot capabilities, though their effectiveness is constrained by the availability of sufficiently rich graph data. Fine-tuning adapts these pre-trained models to specific tasks with strong performance when labels are abundant, but typically requires considerable computational and memory resources, motivating the development of parameter-efficient variants such as GraphLoRA. Graph prompt tuning provides a lightweight alternative by avoiding full model updates and performing well in low-resource or few-shot settings, yet it often underperforms fine-tuning in high-data regimes and struggles with zero-shot transfer without auxiliary techniques. Taken together, these approaches highlight a fundamental trade-off between generalization capacity and efficiency: while Transformers and GFMs enable broad adaptability at high computational cost, prompt-based and adapter-based methods favor scalability and resource efficiency, underscoring the need for adaptive orchestration strategies that align method choice with data availability, task requirements, and computational budgets.

\subsection{Real-world Applications}
\label{subsec:3.4application}

Building on LLM4TAG, recent works address three levels of prediction, namely node, edge, and graph, by pairing text features that are processed by LLMs with graph representations.
At the node level, textual cues enrich local neighbourhoods to classify or rank individual vertices such as atoms in molecules, users in social networks, or products in e-commerce catalogues.
At the edge level, sentence-scale evidence sharpens relational scores such as drug–disease links or user–item affinities, whereas at the graph level global structure pooled with document-level text characterises whole molecules, catalysts, or communities.

\textbf{Node. }
In bioinformatics, MoleculeSTM \cite{liu2023multi} contrasts language-model embeddings of textual descriptions with GNN embeddings of atomic graphs so that every atom inherits rich textual semantics when molecular properties are ultimately predicted. MMF \cite{srinivas2024cross} takes a similar stance: Chebyshev Graph Convolutions generate node embeddings that are cross-attended with text features obtained from zero-/few-shot prompting of an LLM, while a Mixture-of-Experts head dynamically re-weights these embeddings to yield robust per-atom property scores.
Across social platforms, node-centric objectives include user categorization or profiling; here, TAG frameworks \cite{liu2024can,su2024bridging,qiao2024login,zhu2024graphclip} treat profile texts or timelines as node attributes that are blended with neighbourhood signals. In e-commerce, PP-GLAM \cite{choudhary2024interpretable} ensembles language-model outputs with behavioural GNN features so that each product or user node receives an interpretable label, while Shapley additive explanations quantify the textual or structural evidence behind the prediction.

\textbf{Edge. }
In drug discovery, LLM-DDA \cite{gu2024empowering} attaches textual knowledge to drug and disease vertices and then runs a GNN whose message passing is modulated by that text, thereby boosting drug–disease association (edge) prediction. 
EdgePrompt extends a frozen GNN by learning small, trainable vectors on every edge; aggregated during message passing, these prompts inject task-specific text signals without touching the backbone weights.
For chemistry, ChemCrow \cite{bran2023chemcrow} and the CLIP-based approach of \cite{ock2024multimodal} integrate language models with graph encoders so that adsorption-energy regression becomes an edge-aware text-graph matching problem, which is crucial when ranking active–site interactions in catalyst screening.
In social media TAGs, edges often denote replies, follows, or messages; malicious-actor detection frameworks such as \cite{cai2024lmbot,huang2024cgnn} inject post content (edge text) into GNNs to flag suspicious communication links. E-commerce systems routinely perform user–item link prediction for recommendation \cite{roumeliotis2024llms,fang2024gaugllm,wei2024llmrec}, where reviews or product titles serve as edge or contextual text to refine interaction scores.

\textbf{Graph. }
BioBGT \cite{peng2025biologically} is a novel transformer architecture designed specifically for brain graphs, integrating network-entanglement based node importance encoding to capture hub-driven global communication and module-aware self-attention to preserve the brain’s functional segregation and integration.
DrugChat \cite{liang2023drugchat} provides an interactive environment where an LLM reasons over whole-molecule graphs supplied by a GNN and can explain graph-level pharmacological properties. MoleculeSTM, besides its node-level benefits, learns a unified representation that is pooled to predict whole-molecule attributes, while MMF and the graph-assisted pre-training framework of \cite{ock2024multimodal} align entire graph embeddings with sentence-level captions, yielding state-of-the-art performance on molecular property benchmarks.

In social analysis, overlapping-community detection \cite{shchur2019overlapping} treats each community sub-graph as a unit whose textual description (hashtags, topic words) is fused with structural cues to identify latent groups. For recommender systems, bundle recommendation and product understanding tasks \cite{luo2021alicoco2,zhang2022oa,chen2023label} regard a bundle graph as the prediction target, blending review corpora with GNN-derived structural summaries. Catalyst design likewise benefits: the CLIP-style graph–text alignment in \cite{ock2024multimodal} delivers graph-level adsorption-energy estimates that accelerate materials discovery.
Rep-CodeGen \cite{huang2025repcodegen} automatically writes crystal-graph code that satisfies all six key symmetry and continuity constraints, slotting directly into high-throughput pipelines to produce state-of-the-art property predictions without manual descriptor engineering.  
By letting LLM agents continuously adapt representations as new physical rules appear, the framework turns materials screening into a self-optimising, end-to-end process that accelerates discovery at million-scale.
HIGHT \cite{chen2025hight} introduces a hierarchical tokenizer that turns molecules into atom-, motif-, and whole-graph tokens so large language models can reason over molecular graphs with far less hallucination and stronger performance on downstream chemistry tasks.
Llamole \cite{liu2025llamole} is a novel multimodal large-language model designed for inverse molecular design and retrosynthetic planning. By augmenting a base autoregressive LLM with a Graph Diffusion Transformer for multi-conditional molecule generation and a GNN-based reaction predictor for one-step retrosynthesis.

\subsection{Observations and Insights}
\label{subsec:3.5obser}

In this subsection, we summarize the experimental observations and key insights related to LLM for TAG models reported in recent papers.
Specifically, we focus on why LLMs advance graph reasoning and representation learning, as well as the limitations and challenges of designing LLMs for TAG models.

\subsubsection{Why LLMs advance graph reasoning and representation learning?}
\hspace{10pt}

\noindent
We provide the key ideas about how and why LLMs enhance graph reasoning and representation learning. 


\noindent \textbf{Preliminary but promising graph reasoning abilities.}
Several studies \cite{wang2024can, han2023pive, guo2023gpt4graph, zhang2023llm4dyg, luo2023reasoning, huang2023can, zhao2023graphtext, fatemi2023talk, ye2023natural} find that LLMs exhibit preliminary capacity for graph reasoning, particularly under simpler tasks such as cycle detection or shortest path. These works suggest that, even without extensive architectural modifications, LLMs can recognize and reason about graph-structured data when properly prompted.

\noindent \textbf{Enriched attributes and contextual information.}
Multiple research efforts \cite{chen2024exploring, qin2023disentangled, he2023harnessing, su2024bridging} highlight that LLMs can enrich node attributes or refine node embeddings, leading to improvements in downstream tasks ($e.g.$, node classification). 
For instance, generating high-quality textual descriptions for node features or using deep sentence embeddings to augment GNNs has demonstrated both effectiveness and efficiency, showing the synergy between LLM-based text understanding and graph structural modeling.

\noindent \textbf{Enabling implicit graph reasoning.}
Prior studies on multi-hop QA, knowledge probing, or structured commonsense reasoning \cite{ding2023knowledge, he2024g, creswell2022selection} demonstrate that LLMs implicitly learn connections among a vast network of entities during pretraining. 
This suggests that LLMs can tap into their internal ``knowledge graph'' to tackle tasks requiring relational or multi-hop reasoning without explicit GNN components.

\noindent \textbf{Potential viability for addressing complex graph tasks.}
GPT-4–based experiments \cite{yao2024exploring} show that large models can produce rule-based graphs ($e.g.$, trees, cycles, certain regular graphs) and even molecular structures under specific prompts. These findings reveal a capacity for generative modeling of graph structures, opening up avenues for data augmentation or synthetic graph creation.
In addition, Zhao et al. \cite{zhao2024dynllm} clarified that LLMs have preliminary spatial-temporal understanding abilities on dynamic graphs.

\noindent \textbf{Applicability across diverse domains and modalities.}
By virtue of their strong language-based feature extraction and knowledge embedding capabilities, LLMs have been explored in multi-modal or domain-specific scenarios, such as drug discovery or biomedical ontologies \cite{chen2024exploring, li2024chain, huang2024can, dai2024large, li2024empowering}. 
This indicates an emerging potential for LLMs to unify textual and graph-structured information under a single large model paradigm.

\noindent \textbf{Effective structural information usage through tailored prompts.}
Studies like \cite{huang2024can} and \cite{guo2023gpt4graph} demonstrate that structural prompts, including neighborhood homophily encoding and role-based designs, allow LLMs to process graph structure as linearized text while achieving competitive performance on structural reasoning tasks. Techniques like natural language prompts encoding multi-hop connections \cite{ye2023natural} and adjacency-based chain-of-thought prompting \cite{fatemi2023talk} further validate the ability of LLMs to bridge the gap between textual and structural representations.

\subsubsection{What are the limitations and challenges of integrating LLMs with TAGs learning?}
\hspace{10pt}

\noindent
We also discuss the key insights summarized from existing works about the limitations and challenges of integrating LLMs with TAG learning.


\noindent \textbf{LLMs struggle with explicit topology representation.}
    LLMs process graph structure as contextual text rather than explicit representations, leading to limited performance in topology-dependent tasks like causal inference and multi-hop reasoning \cite{chen2024clear, guo2023gpt4graph, guan2025attention}. This textual interpretation constrains their ability to generalize to complex graph tasks.

\noindent \textbf{LLMs exhibit brittleness to spurious correlations and a tendency toward in-distribution memorization.}
    Works \cite{wang2022self, fatemi2023talk}  note that LLMs can latch onto spurious or superficial correlations in graph benchmarks, indicating they may be memorizing in-distribution patterns rather than genuinely reasoning. Similarly, the  NLGIFT \cite{zhang2024can} study underscores that LLMs often fail to generalize out-of-distribution when the synthetic training data’s patterns shift.

\noindent \textbf{LLMs show diminishing returns when applied to complex graph tasks.}
    Multiple works \cite{jin2024large, pan2024unifying, zhou2022least, chen2024exploring, wang2024can} show that while advanced prompting methods ($e.g.$, chain-of-thought, least-to-most, self-consistency) help on simpler tasks, the benefits fade on more intricate graph problems such as topological sorting or Hamiltonian path. This suggests a gap between LLMs’ naive text-driven reasoning and the rigorous algorithmic frameworks needed for complex graph queries.

\noindent \textbf{LLMs exhibit a lack of robustness when processing large or dense graphs.}
    Studies on dynamic graphs \cite{zhao2024dynllm} and dense structures report sharp performance declines as graph size and density grow. 
    LLMs may handle small synthetic graphs but struggle with scalability—both in memory and in effectively interpreting complex structural cues from large adjacency matrices or highly connected topologies.

\noindent \textbf{LLMs face ambiguities in distinguishing and effectively leveraging graph structures compared to textual context.}
    Recent works \cite{huang2024can, chen2024exploring, guo2023gpt4graph} indicate that LLMs frequently treat graph prompts as unstructured paragraphs, ignoring explicit topological patterns. They focus on textual overlap or keywords rather than genuinely parsing adjacency relations. 
    Hence, even when structural prompts are provided, the model might predominantly rely on contextual cues instead of actual graph structure. 

\noindent \textbf{Potential for Inaccurate Outputs and Data Leakage.}
     Studies \cite{chen2024graphwiz, tang2024higpt, chen2024exploring, pan2024unifying} indicate that some LLM-as-Predictor approaches face uncontrollable outputs ($e.g.$, invalid labels or hallucinated edges) and potential test data leakage due to large pretraining corpora. These limitations undermine trustworthiness, especially in high-stakes tasks where correctness and structure compliance are crucial ($e.g.$, molecule generation).

%% file: s5-tag2llm.tex
\section{TAG to LLMs}
\label{sec:tag2llm}

\begin{figure}
    \centering
    \includegraphics[width=0.9\linewidth]{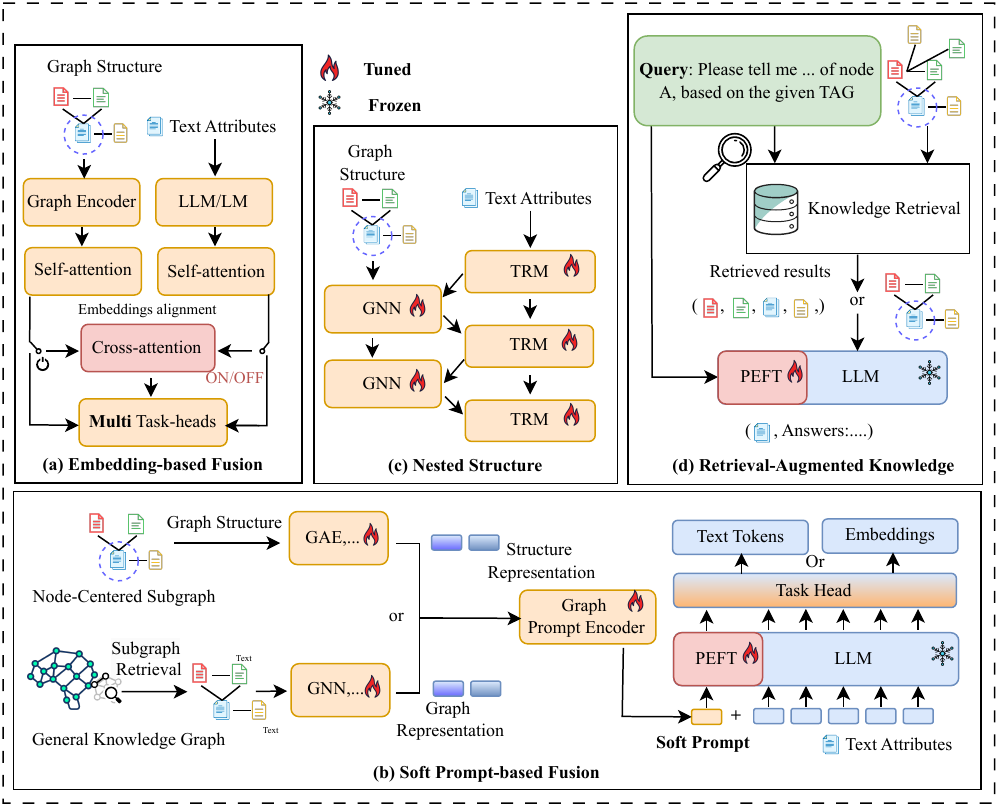}
    \caption{The illustration of  orchestration approaches of TAG4LLM, which contain (a) Embedding-based fusion, (b) Soft prompt-based fusion, (c) Nested structure, and (d) Retrieval-augmented knowledge.}    \label{fig:tag2llm}
\end{figure}

Unlike the LLM4TAG paradigm, which applies existing LLM techniques to TAG tasks, TAG4LLM reverses this process, as shown in \Cref{fig:tag2llm}. It first builds and leverages the intrinsic TAG topology to enrich and steer the LLM, thereby enhancing its reasoning capabilities and factual accuracy.
Existing studies can be grouped into two orchestration styles: 
\textbf{(1) Two-module orchestration:} a separate graph encoder, typically a GNN, independently generates topology aware embeddings or prompt vectors; these representations are then supplied to LLMs, preserving the modular separation of the two components.
\textbf{(2) Multi-module orchestration:} graph transformer reasoning is woven directly into language model layers or task specific subgraphs and verbalized triples are retrieved on demand to ground the decoder.
Existing works of both styles demonstrate that explicit relational cues from TAGs lessen hallucinations, enable multi-hop reasoning, and raise the factual accuracy of LLMs.

\subsection{Two-Module Orchestration}
\label{4.1}
Two-model orchestration refers to a paradigm in which graph and language components are trained separately but combined via lightweight fusion mechanisms.
Specifically, \textbf{(1) Embedding-based fusion} first lets a GNN encode the TAGs, then linearly (or via a cross-modal adapter) projects the resulting graph embeddings into tokens or prefix vectors that are concatenated with the LLM’s input, giving the model direct access to graph structure (\Cref{fig:tag2llm}a).
\textbf{(2) Soft-prompt-based fusion} instead converts the graph output into a small set of trainable prompt embeddings placed before the textual prompt; only these prompt vectors are updated, yielding a parameter-efficient way to condition the frozen LLM (\Cref{fig:tag2llm}b).

\subsubsection{Embedding-based Fusion}
\hspace{10pt}
\label{subsec:embedLLM}

\noindent
Integrating GNN-processed embeddings from TAGs with LLMs enhances reasoning by leveraging the inherent graph structure. GNNs generate structure-aware embeddings \( \mathbf{H} = f_{\text{GNN}}(\mathbf{X}, \mathbf{A}) \), where \( \mathbf{X} \) is the node embedding matrix and \( \mathbf{A} \) is the adjacency matrix. These embeddings are then passed into LLMs for prediction: \( \tilde{\mathbf{Y}} = \text{Parse}(f_{\text{LLM}}(\mathbf{H}, p)) \), where \( p \) represents any additional parameters or prompts required by the LLM to generate the output. This integration improves the model’s structural understanding but often requires tuning to align the prediction outputs with the desired format.
Sharing the similiar ideas, in \Cref{subsub:softPrompt}, we will review several works on soft prompts, which are learnable embeddings generated by prompt encoders, such as GNNs, that act as a bridge between TAGs and LLMs, enabling their integration for improved LLM reasoning.

Based on the embedding-based fusion ideas, MolCA \cite{liu2023molca} leverage BLIP-2’s QFormer \cite{li2023blip} as a cross-modal projector, which maps the output of the graph encoder to the input text space of LLMs. 
LLaGA \cite{chen2024llaga} pre-trains model by fusing both stages of BLIP-2 together to obtain downstream results.
GraphLLM \cite{chai2023graphllm} enhances the LLM’s reasoning by generating a graph-augmented prefix through linear projection of the graph representation during prefix tuning, enabling the LLM to integrate structural information from the graph transformer. 
 In contrast, GraphGPT \cite{tang2024graphgpt} and InstructMol \cite{cao2023instructmol} utilize a simpler linear layer to project the encoded graph representation into graph tokens, which are then aligned with textual information by the LLM, facilitating seamless integration of graph and text data.

\subsubsection{Soft Prompt-based Fusion}
\label{subsub:softPrompt}
\hspace{10pt}

\noindent
Soft prompts \cite{lester2021power} are trainable, continuous embeddings designed to efficiently adapt pre-trained language models ($e.g.,$ GPT-4) for downstream tasks while maintaining scalability by keeping the core model parameters, $\theta$, fixed. This approach can be formalized as $\text{Pr}_{\theta;\theta_p}(\mathcal{Y} \mid [\mathcal{P}_\text{soft}; \mathcal{S}])$, where $\mathcal{P}_\text{soft}$ represents the learnable soft prompt embeddings generated by a designed prompt encoder, and $\mathcal{S}$ denotes task-specific tokens. 
The conditional generation process is optimized by maximizing the likelihood of $\mathcal{Y}$ through backpropagation, with gradient updates applied exclusively to $\theta_p$.

In the context of TAGs, soft prompts act as a bridge between TAGs and LLMs, enabling seamless integration of graph structural information with the textual capabilities of LLMs.
Existing works on soft prompting primarily focus on the initialization mechanism of $\mathcal{P}_\text{soft}$, emphasizing the incorporation of graph topological information through graph projectors/adapters such as GCN to enhance performance. 
Input text attributes $\mathcal{S}$ can range from raw text, pre-processed tokens via LMs, or text embeddings (potentially adapted with auxiliary modules) to graph descriptions and task-specific queries.

GraphAdapter \cite{huang2024can} employs GNNs as adapters to produce soft prompts that integrate graph structure into frozen LLMs, aligning structural and contextual representations through a fusion mechanism. 
By leveraging prompt-aware fine-tuning, GraphAdapter transforms tasks into next-token prediction, enabling efficient adaptation of structural information for downstream TAG tasks. 
Sharing a similar concept with GraphAdapter,  LLaGA \cite{chen2024llaga} and GALLM \cite{luo2024enhance} preserve the general-purpose capabilities of LLMs while adapting graph data into a format compatible with LLM inputs. 
It accomplishes this by reorganizing graph nodes into structure-aware sequences and subsequently mapping these sequences into the token embedding space using a versatile projector.
To be more efficient, GPEFT \cite{zhu2024parameter} and NT-LLM \cite{ji2024nt} further adopt Parameter-Efficient Fine-Tuning (PEFT) \cite{mangrulkar2022peft}, integrating modules like LoRA \cite{hu2021lora} and Prefix-Tuning \cite{li2021prefix} to adapt models with minimal overhead.
Further more, recent works like G-Retriever \cite{he2024g}, DRAGON \cite{yasunaga2022deep}, SubgraphRAG \cite{li2024simple}, AskGNN \cite{hu2024let}, and GNP \cite{tian2024graph} share a common emphasis on integrating retrieval mechanisms to enhance soft prompt quality, which will be discussed more in \Cref{subsec:RAG}.
\subsubsection{Discussion.}
\label{subsec:disciss41}
\hspace{10px}

Two-module orchestration demonstrates how graph and language components can remain decoupled yet cooperate through lightweight fusion. Embedding-based fusion directly injects graph encoder outputs into LLM input space, offering a straightforward way to expose structural signals but often requiring careful alignment and potentially larger adaptation layers. Soft-prompt-based fusion, by contrast, focuses on parameter efficiency: graph-aware prompts act as small trainable vectors that condition a frozen LLM, preserving general-purpose capabilities while selectively injecting graph structure. Both schemes share the idea of treating graph encoders as front-end adapters to LLMs, yet differ in their trade-offs. Embedding-based methods typically achieve stronger coupling but incur higher adaptation cost, whereas soft prompts scale more easily across tasks but can struggle to match fine-tuned performance in high-data regimes. Together, these designs highlight a spectrum of efficiency–expressiveness trade-offs in orchestrating TAGs and LLMs.

\subsection{Multi-Module Orchestration}
\label{4.2}

Multi-model orchestration folds graph reasoning directly into the language model and can be broadly categorized into two schemes. \textbf{(1) Nested structures} embed a graph transformer inside the LLM, interleaving message-passing layers with self-attention so that structural signals and textual semantics are refined in tandem (\Cref{fig:tag2llm}c). \textbf{(2) Retrieval-augmented knowledge} pairs the LLM with a dynamic retriever that supplies task-specific subgraphs or verbalized triples, which are encoded and fed back as additional context to ground generation (\Cref{fig:tag2llm}d).

\subsubsection{Nested Structure}
\hspace{10pt}
\label{subsec:nested}

\noindent
The graph-nested structure (\Cref{fig:tag2llm}c), as exemplified by Graphformer \cite{yang2021graphformers}, dynamically aligns graph topology with textual semantics by embedding GNN reasoning within transformer layers. Node embeddings, derived from token representations such as the \texttt{[CLS]} token, are processed through a graph transformer to capture relational and structural patterns. 
The resulting graph-transformer output is concatenated with the input embeddings of the transformer layer, allowing iterative refinement of graph-structured and semantic information across layers.
Codeformer \cite{liu2023codeformer} employs a tightly coupled, iterative loop that alternates Transformer-based multi-head attention for intra–basic-block token encoding with GRU-gated message passing and aggregation over the control-flow graph, so node and graph embeddings are co-refined in a single end-to-end pass.
Flat Tree-based Transformer \cite{mao2025flat} reframes nested named entity recognition (NER) as a joint text-and-graph problem: token embeddings from BERT are merged into span representations, but only those spans that match nodes in the constituency-parse tree are kept, and the tree’s edges supply a local attention mask while a parallel global mask preserves sentence-level context.
Gao et al. \cite{gao2025llm} introduces a nested architecture that positions hierarchical graph encoders inside a large language model enhancer, letting token-level semantics flow outward while neighborhood-level cues flow inward. Treating this arrangement as a causal pipeline, attention bridges dynamically align language signals with graph substructures, improving interpretability and task accuracy.
\begin{figure}
    \centering
    \includegraphics[width=1.0\linewidth]{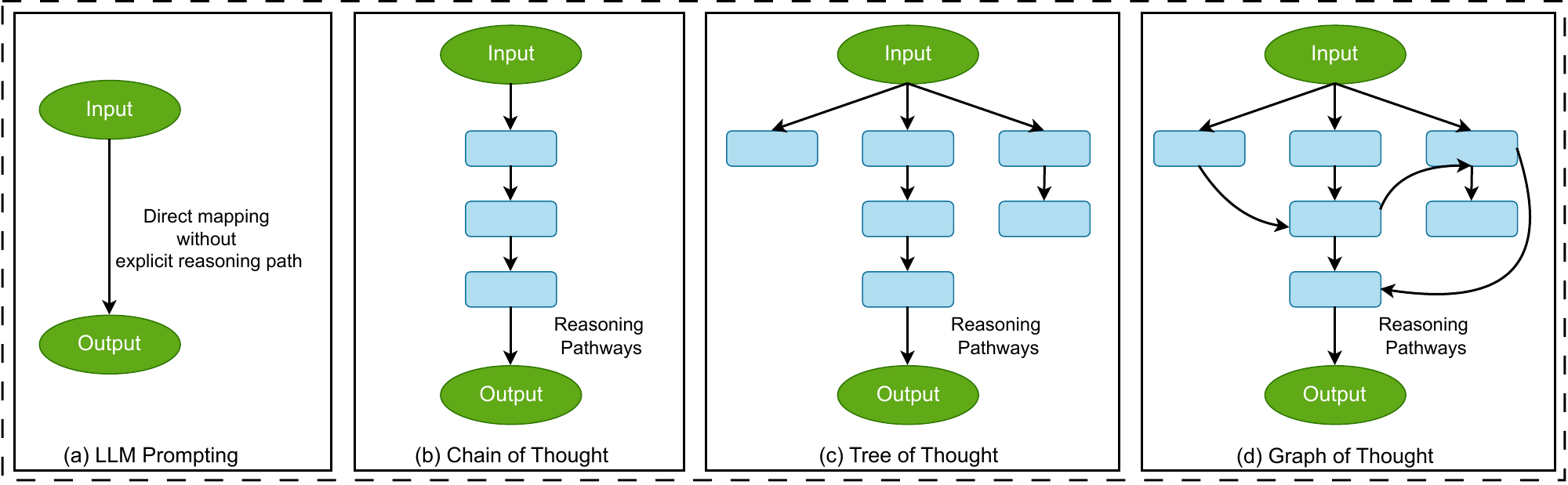}
    \caption{The illustration of orchestration approaches of TAG4LLM related to the graph of thought.}    \label{fig:tog}
\end{figure}

\subsubsection{Retrieval-Augmented Knowledge}
\hspace{10pt}
\label{subsec:RAG}

\noindent
Retrieval-augmented knowledge, generated from TAGs for LLMs, integrates retrieval mechanisms \cite{gao2023retrieval} with graph-based representations to enhance the reasoning and contextual understanding capabilities of LLMs.
In this framework, external information from sources such as knowledge graphs, text corpora, or structured databases is retrieved and organized as \textbf{(1)} sub-graphs of input TAGs or \textbf{(2)} text tokens.
These results are subsequently integrated with LLMs to provide enriched, structured context for downstream tasks. 
On the one hand, GNNs or similar graph models are commonly employed to encode the retrieved graph's structure, generating representations that seamlessly integrate with the LLM's input embeddings. 
Techniques outlined in \Cref{subsub:softPrompt} can be applied to enhance this integration.
On the other hand, retrieved unstructured text tokens can be integrated as supplementary context into Seq2Seq LLMs to enrich and enhance the generation process. 
Works \cite{wang2022self, fatemi2023talk}  note that LLMs can latch onto spurious or superficial correlations in graph benchmarks, indicating they may be memorizing in-distribution patterns rather than genuinely reasoning. Similarly, the  NLGIFT \cite{zhang2024can} study underscores that LLMs often fail to generalize out-of-distribution when the synthetic training data’s patterns shift and retrieval-augmented knowledge is helpful.

\noindent
\textbf{Based on retrieved sub-graphs.}
G-Retriever \cite{he2024g} is an representative framework designed to improve retrieval and reasoning over graph-structured data through four stages: \textit{(1) Indexing}, which organizes and indexes graphs for efficient query processing; \textit{(2) Retrieval}, where relevant nodes and edges are retrieved based on the query; \textit{(3) Sub-graph Construction}, which extracts a connected sub-graph with as many relevant nodes and edges as possible while maintaining a manageable size; and \textit{(4) Generation}, where a response is generated using a ``graph embedding,'' a textualized version of the subgraph combined with the query, enabling smooth integration with downstream language models.

Sharing similar ideas, REALM \cite{guu2020retrieval} emphasizes a knowledge retriever that enables the model to retrieve and attend to relevant documents from a large corpus during pre-training, significantly enhancing performance in open-domain question answering.
In contrast, DRAGON \cite{yasunaga2022deep} focuses on a cross-modal module to deeply integrate text and knowledge graph (KG) modalities, aligning text segments with relevant KG subgraphs to produce fused token and node representations for enhanced reasoning and representation learning. 
AskGNN \cite{hu2024let} integrates a GNN-powered retriever to select graph examples, employing soft prompts to incorporate task-specific signals and enhance LLM performance in node-centric tasks. 
Furthermore, GNP \cite{tian2024graph} extends this paradigm with graph neural prompting, leveraging GNN encoders, cross-modality pooling, and domain projectors to augment LLMs with structured graph knowledge, excelling in commonsense and biomedical reasoning.
$R^{2}$-Guard \cite{kang2025r2guard} enhances LLM inference by combining data-driven category-specific unsafety predictions with explicit logical reasoning encoded as first-order rules in probabilistic graphical models, thus capturing and leveraging interdependencies among safety categories

\noindent
\textbf{Based on retrieved text tokens.}
Mindmap \cite{wen2023mindmap}, ChatRule \cite{luo2023chatrule}, and TripletRetrieval~\cite{li2023graph} explore methods to adapt knowledge graphs (KGs) for reasoning tasks with LLMs by leveraging structured-to-text transformations and logical reasoning. 
TripletRetrieval and Mindmap share the idea of converting graph structures into text representations for LLMs: TripletRetrieval utilizes pre-defined templates to transform triples into short sentences, while Mindmap organizes graph structures into mind maps that consolidate KG facts and LLMs' implicit knowledge for reasoning. 
In contrast, while ChatRule also relies on verbalization like Li et al. and Mindmap, ChatRule adopts a different approach, sampling relation paths from KGs, verbalizing them, and prompting LLMs to generate logical rules for reasoning.
What's more, GCR \cite{luo2025gcr} turns knowledge-graph paths into a trie that hard-constrains an LLM’s token generation, letting a small KG-specialized model explore faithful paths while a stronger general LLM fuses them—achieving hallucination-free, state-of-the-art KGQA that even zero-shots to unseen graphs.
By injecting graph structure directly into the decoding loop, the work charts a scalable recipe for future graph-augmented language systems that balance efficiency, accuracy, and cross-KG generalization.


\subsubsection{Graph of Thought.}
\label{subsec:tog}
\hspace{10px}

\noindent
Recent work replaces the linear chain of thought \cite{wei2022chain} and the tree of thought \cite{yao2023tree} with graph-based reasoning that aligns naturally with knowledge graphs and TAGs. 
As illustrated in \Cref{fig:tog}, Graph of Thoughts \cite{besta2024graph} extends chain- and tree-style reasoning by representing each intermediate idea as a node and connecting related ideas with edges. Instead of following one fixed path, the model can generate several candidate thoughts in parallel, connect them into a graph, and then select the most promising subgraph for the final answer. This design allows information to be reused across branches and supports backtracking when an earlier step is unreliable. 
Think on Graph \cite{sun2023think} grounds each reasoning step on an external knowledge graph by iteratively traversing entities and relations under LLM guidance, which reduces hallucination and supports multi hop factual inference. 
What is more, task focused adaptations follow the same principle while tailoring node and edge semantics to data domains, including {ReX-GoT} \cite{zheng2024reverse} for multi choice dialogue commonsense that iteratively excludes distractors on an option centric thought graph, {GOT4Rec} \cite{long2024got4rec} for sequential recommendation that wires user intent, item semantics, and temporal cues into a unified thought graph, and {Thought Graph \cite{hsu2024thought}} for biological discovery that aligns thought nodes with ontology concepts and curated relations to compose mechanistic hypotheses. 
A synthetic perspective \cite{besta2025demystifying} compares chains, trees, and graphs and argues that graph structures uniquely support cyclic evidence integration, cross path reuse, and non monotonic revision while remaining compatible with retrieval and agentic planning. 
Collectively these frameworks share the idea of externalizing intermediate reasoning as nodes and edges so that LLMs can retrieve, compose, verify, and revise evidence on KGs and TAGs, yet they differ in how structure is supplied, free form within the model for Graph of Thoughts, grounded by external knowledge graphs for Think on Graph and MindMap, or specialized for a target domain as in ReX-GoT, GOT4Rec, and Thought Graph.

\subsubsection{Discussion.}
\label{subsec:disciss42}
\hspace{10px}

\noindent
Multi-module orchestration highlights two complementary pathways for coupling graphs with LLMs. Nested structures pursue \emph{tight integration}, embedding graph reasoning mechanisms directly into transformer layers so that structural signals and textual semantics are refined in tandem. This design enables fine-grained alignment of topology and language but often requires substantial architectural modification and training cost. 
In contrast, retrieval-augmented knowledge adopts a more \emph{modular interface}, pairing LLMs with external retrievers that supply task-relevant subgraphs or verbalized triples. This approach emphasizes scalability and flexibility: the LLM remains largely unchanged while retrieved graph evidence grounds its predictions, though effectiveness depends heavily on retrieval quality and the efficiency of subgraph construction.

Thought-on-Graph frameworks can be viewed as a natural extension within this orchestration spectrum. They externalize intermediate reasoning steps as graph nodes and edges, enabling reuse, revision, and parallel exploration of evidence that nested or retrieval-based designs alone cannot fully capture. Together, these methods reveal a trade-off between integration depth and system modularity. Nested designs maximize representational synergy but are costly to adapt across domains, retrieval-augmented approaches scale broadly but risk shallow coupling, and graph-based thought structures offer a flexible middle ground that foregrounds reasoning interpretability. Future research may benefit from hybrid orchestration strategies that adaptively switch between these schemes depending on task complexity, data availability, and computational budget.

\subsection{Real-world Applications}
\label{4.3applicaiton}
TAG2LLM applications enhance textual knowledge from LLMs with graph information at three complementary granularities. 
At the node level, they enrich each node’s neighbourhood with language-level semantics to rank or classify individual entities. 
At the edge level, they combine sentence-scale evidence with relation-aware message passing to predict or re-score specific links. 
At the graph level, they pool global structure with document-level text so the model can label or describe an entire graph in a single shot. 

\textbf{Node. }Node‐level applications label or rank individual vertices by weaving textual semantics into local graph structure.  In clinical NLP, Yue et al. \cite{yue2020graph} build a term–term co-occurrence graph from millions of electronic medical records, then classify the semantic type of each medical term node with graph-aware learning.  SPECTRA \cite{ektefaie2024evaluating} treats molecular sequences as nodes linked by shared spectral properties and predicts sequence-level phenotypes such as antibiotic resistance, while code-understanding systems encode functions or variables as textual nodes to improve summarisation, naming, and bug detection \cite{zhang2024galla,vadoce2024enhancing}.

\textbf{Edge. }Edge‐level applications focus on predicting or auditing relations by combining sentence-scale evidence with edge-aware reasoning.  Scorpius \cite{yang2024poisoning} shows how large language models can fabricate abstracts that insert false drug–disease links into medical knowledge graphs, exposing the vulnerability of such edges.  In astrophysics, the two-stage model of \cite{shaoastronomical} joins textual embeddings with multi-relational graph structure to boost cross-catalogue link prediction between celestial objects, and BIORAG \cite{wang2024biorag} uses hierarchical retrieval over twenty-two million papers to supply accurate edge information for life-science question answering.

\textbf{Graph. }Graph‐level applications assign a holistic label or generate content for an entire graph by pooling global structure with corpus-level text.  ATLANTIC \cite{munikoti2023atlantic} incorporates structural relationships among interdisciplinary scientific documents into a retrieval-augmented language model, improving scientific question answering and document classification.  These graph-wide techniques demonstrate how the fusion of text and topology can advance large-scale reasoning tasks that transcend individual nodes or edges.
The study MVQA \cite{baovisual} builds HIE Reasoning, a multimodal benchmark that joins neonatal brain MRI with clinician-verified questions to probe professional level reasoning by vision language models. It also presents Clinical Graph of Thought, a graph structured prompting strategy that fuses visual cues and textual clinical knowledge to follow expert diagnostic steps and markedly improves prediction of two year neurocognitive outcomes.
Among the growing set of knowledge-graph–enhanced LLM applications, K-RagRec \cite{wang2025knowledge} demonstrates how structured retrieval can strengthen recommendation.  It indexes multi-hop neighborhoods with a GNN, selectively retrieves and re-ranks the most relevant subgraphs, and projects their embeddings into the LLM’s prompt, thereby reducing hallucinations and delivering more accurate, up-to-date recommendations.

\subsection{Observations and Insights}
\label{4.4obser}

In this subsection, we introduce the empirical observations and insights of TAG for LLM techniques.
Specifically, we summarize the insights and provide some future directions for the synergy between LLM and TAGs. 


\noindent \textbf{Enhancing generalization of LLMs beyond pattern memorization.}
The NLGIFT benchmark highlights the need for robust out-of-distribution testing and post-training alignment to push LLMs beyond memorized patterns. Future methods could incorporate novel finetuning strategies that disentangle superficial correlations from true relational reasoning, helping LLMs adapt to real-world, evolving graph tasks.

\noindent \textbf{Improved Scalability via modular architectures or retrieval-based interfaces.}
Several works \cite{he2024g, li2024simple, dai2025graphpattern,zhao2024dynllm} emphasize the importance of modular pipelines that retrieve relevant subgraphs, letting the LLM focus on a manageable local context. This approach could mitigate the overhead of prompting massive graphs in full, and allow efficient integration of structure and textual knowledge.

\noindent \textbf{The development of refined benchmarks and evaluation protocols.}
Various newly proposed benchmarks—NLGraph \cite{wang2024can}, GPT4Graph \cite{guo2023gpt4graph}, NLGIFT \cite{zhang2024can}—demonstrate the community’s efforts to rigorously evaluate LLM-based graph reasoning. 
Future developments in LLM-based graph learning necessitate multi-faceted tasks ($e.g.$, subgraph matching, graph similarity search) and comprehensive metrics (correctness, interpretability, robustness, efficiency), ensuring that ongoing research can reliably assess and drive genuine progress in this emerging area.

\noindent \textbf{Beyond textual parsing and focus on algorithmic integration.}
It is crucial to move beyond textual parsing and focus on algorithmic integration, enabling LLMs to effectively incorporate structural and algorithmic reasoning for graph tasks.
Some theoretical findings \cite{feng2024towards, wucan} suggest LLMs might simulate graph algorithms via chain-of-thought. However, purely text-driven approaches are often suboptimal. Incorporating symbolic or programmatic modules—for instance, letting the LLM generate code that executes partial graph algorithms—could yield more robust and interpretable solutions \cite{madaan2022language, sclar2023minding, jin2023patton}.

\noindent \textbf{Expanding the scope of graph tasks and conducting deeper and more comprehensive evaluations.}
Beyond node-level classification and link prediction, more challenging and diverse graph problems, such as subgraph matching, graph similarity search, and motif detection should be incorporated into future benchmarks \cite{li2024glbench}. 
These tasks often require explicit structural traversal, deeper topological matching, or quantitative measures of graph similarity, providing a more thorough test of LLMs’ capacity to handle real-world complexities in graph-based learning.

%% file: s6-direction.tex
\section{OPPORTUNITIES AND FUTURE DIRECTIONS}
\label{sec:futureDire}
This section provides emerging opportunities and articulates six key research directions for advancing text-attributed graph learning: 
large text-attributed graph model, autonomous agents, TAGs in black-box LLM inference, TAG data management, {efficiency and scalability}, and expanding graph task diversity.

\noindent
\textbf{TAG Foundation Models.} Developing foundation models for large scale TAGs requires the same appetite for data and capacity for scale as GPT style language models. 
Pre-training should ingest massive text corpora together with richly connected graph topologies of comparable size. 
A TAG foundation model with billions of parameters can generalize across domains such as social network analysis, e-commerce recommendation, knowledge graph completion and entity resolution while supporting a diverse suite of downstream tasks including node classification, link prediction, graph guided question answering and multi hop reasoning.
Specifically, TAGs share common graph properties ($e.g.$, small-world structure) and linguistic features ($e.g.$, semantic similarity, contextual embeddings), which can enhance generalization. 
The core idea is to train models that capture both domain-agnostic representations and domain-specific nuances through modular architectures or meta-learning. 
For multi-task learning, task-aware mechanisms, such as dedicated heads or attention layers, focus on specific objectives, while advanced techniques like loss weighting and gradient normalization optimize learning across tasks.
Although we have discussed several preliminary approaches in \Cref{subsec:pretrainTAG} that offer valuable insights, they remain narrowly focused on individual technical challenges such as unifying domain representations, aligning graph feature dimensions, and harmonizing downstream task objectives. 
To date, a genuinely foundational model for TAGs continue to face numerous challenges and require further exploration and development.

\noindent
\textbf{Autonomous Agents.}
LLM-based autonomous agents \cite{schick2024toolformer, wang2024survey, xi2023rise} leverage LLMs to independently perform complex tasks through natural language understanding, reasoning, and decision-making. Autonomous multi-agent systems offer a promising solution for TAGs by combining distributed reasoning with task specialization to handle both structural and semantic complexity. For TAGs, which combine graph topology with textual attributes, approaches must balance graph processing with natural language understanding. Multi-agent systems can enhance TAG representation learning by assigning roles: structural agents model topological patterns, while semantic agents extract insights from textual data. Decision-making agents, using reinforcement learning, integrate these insights to optimize tasks like multi-hop reasoning and graph completion. Collaboration among agents, supported by multi-agent reinforcement learning \cite{zhang2021multi} and attention-based communication, is key for scaling to large TAGs.
Moreover, applying graph-learning algorithms to optimize agent routing presents a promising research direction for minimizing latency and communication overhead in distributed AI systems \cite{feng2025graphrouter, ding2024grom}.




\noindent
\textbf{TAGs in Black-Box LLM Inference.}
Integrating knowledge from TAGs into black-box LLMs remains a significant challenge, as models like GPT-4 often restrict access to their internal structures, rendering traditional methods of architectural modification or fusion module integration impractical. Current LLM prompts primarily rely on linear or chain-of-thought progression, which is insufficient to fully capture the structured relationships inherent in TAGs. A promising direction involves enhancing TAGs by constructing relational structures from unstructured text, thereby transforming raw text into TAGs for more effective prompt design and input preparation. 
Moreover, hybrid approaches that combine lightweight external knowledge modules with prompt-based methods could further enhance reasoning capabilities and reliability, especially in complex multi-hop reasoning tasks. Such innovations would bridge the gap between TAGs and black-box LLMs, enabling interpretable, accurate, and scalable inference.

\noindent
\textbf{TAG Data Management.}
TAGs possess rich textual information, posing challenges for existing graph database systems ($e.g.$, Neo4j, ArangoDB\footnote{https://neo4j.com/, https://ongdb.com/}) to effectively handle TAGs by supporting efficient storage, indexing, and querying of text-embedded attributes alongside graph data. 
This may require hybrid systems combining graph databases with vector search engines to manage the computational and storage overhead of advanced text embeddings ($e.g.$, BERT and GPT-4).
Vector databases, such as FAISS \cite{johnson2019billion}, are particularly well-suited for handling high-dimensional embeddings, enabling efficient similarity search and retrieval of text-embedded data. 
These systems could complement graph databases by providing fast access to semantically rich text representations, facilitating advanced queries on TAGs.
Optimizing the performance for high-dimensional text embeddings and graph queries will necessitate integration of verctor data and graph data query engines. 
Addressing these technical challenges will enable TAGs to tackle complex real-world problems such as semantic knowledge extraction and hybrid graph-text analytics.

\noindent
\textbf{Efficiency and Scalability.}
While parameter-efficient adapters such as LoRA \cite{hu2022lora} have already lowered the training‐time memory footprint and compute cost of coordinating large language models with graph encoders, genuine end-to-end scalability is still far from solved.  Real-world deployments increasingly demand that a single pipeline handle graphs with hundreds of millions of vertices while simultaneously hosting multi-billion-parameter LLM backbones—something current prototypes manage only under laboratory conditions.  Future work should therefore investigate lighter, query-adaptive retrieval policies that bypass uninformative subgraphs, sparsity-aware optimization that prunes redundant LLM activations on the fly, and hierarchical caching strategies that pin hot subgraphs in faster memory tiers.  In addition, asynchronous, distributed TAG–LLM execution frameworks could overlap GNN message passing with LLM decoding, exploiting modern GPU clusters and high-bandwidth interconnects to hide latency.  Complementary advances in mixed-precision quantization, parameter sharing, and node-level locality scheduling would further allow trillion-edge graphs to be streamed through billion-parameter models while staying within realistic power and latency budgets.

\noindent
\textbf{Expanding Graph Task Diversity.}
To more rigorously assess LLMs on graph‐based learning, future benchmarks should move beyond node classification and link prediction to include tasks such as subgraph matching, graph similarity search, and motif detection \cite{li2024glbench}. 
These problems demand explicit structural traversal, deeper topological alignment, and quantitative similarity measures, thereby providing a more comprehensive evaluation of an LLM’s ability to navigate and reason over complex, real-world graph structures.
Multiple works \cite{jin2024large, pan2024unifying, zhou2022least, chen2024exploring, wang2024can} show that while advanced prompting methods ($e.g.$, chain-of-thought, least-to-most, self-consistency) help on simpler tasks, the benefits fade on more intricate graph problems such as topological sorting or Hamiltonian path. This suggests a gap between LLMs’ naive text-driven reasoning and the rigorous algorithmic frameworks needed for complex graph queries.
Studies on dynamic graphs \cite{zhao2024dynllm} and dense structures report sharp performance declines as graph size and density grow. 
    LLMs may handle small synthetic graphs but struggle with scalability—both in memory and in effectively interpreting complex structural cues from large adjacency matrices or highly connected topologies.

\section{CONCLUSION}
\label{sec:conclusion}
The integration of large language models (LLMs) with text-attributed graphs (TAGs) is rapidly maturing into a coherent research area. Viewing the literature through an orchestration lens, we synthesized how LLMs enhance TAG learning (LLM4TAG) and how TAGs strengthen LLM reasoning (TAG4LLM) via sequential and parallel pipelines, pre-training and prompt-based adaptations, as well as two-module and multi-module designs including retrieval-augmented and nested architectures. We mapped applications across node, edge, and graph levels, and distilled empirical observations on where these systems excel and where they fall short—most notably in explicit structural reasoning, robustness, scalability, and faithful, interpretable generation.

Beyond unifying techniques and findings, this survey curates resources and articulates forward paths: building TAG foundation models, designing agentic and retrieval-centric frameworks for efficiency, enabling black-box LLM integration, advancing TAG data management, and expanding benchmarks to stress true structural competence. We hope the taxonomy, insights, and resources assembled here provide a practical blueprint for developing transparent, reliable, and scalable TAG–LLM systems, and for accelerating progress toward general, structure-aware language intelligence.
